\documentclass[letterpaper]{article} 
\usepackage{aaai2027}  
\nocopyright
\usepackage[hyphens]{url}  
\usepackage{graphicx} 
\usepackage{natbib}  
\usepackage{caption} 
\usepackage{algorithm}
\usepackage{algorithmic}
\usepackage{amsmath}
\usepackage{amssymb}
\usepackage{multirow}

\usepackage{booktabs}
\usepackage{tabularx}
\newcommand{\MSE}{\operatorname{MSE}}
\newcommand{\sg}{\operatorname{sg}}
\usepackage{array}
\usepackage{colortbl}
\usepackage{newfloat}
\usepackage{listings}
\DeclareCaptionStyle{ruled}{labelfont=normalfont,labelsep=colon,strut=off} 
\floatstyle{ruled}
\newfloat{listing}{tb}{lst}{}
\floatname{listing}{Listing}

\usepackage{booktabs}

\title{AS-FedBridge: Pseudo-Spike Bridge Distillation for Heterogeneous ANN-SNN Federated Learning}
\author{
    Shengyang Li\textsuperscript{\rm 1},
    Yiting Dong\textsuperscript{\rm 2} \corresponding,
    Liuyang Song\textsuperscript{\rm 1},
    Ximing Wang\textsuperscript{\rm 1},
    Luyuan Xie\textsuperscript{\rm 1},
    Cong Li\textsuperscript{\rm 1},
    Qingni Shen\textsuperscript{\rm 1} \corresponding,
    Zhaofei Yu\textsuperscript{\rm 2,\rm 3} 
}
\affiliations{
    \textsuperscript{\rm 1}School of Software and Microelectronics, Peking University, Beijing, China
    \textsuperscript{\rm 2}School of Computer Science, Peking University, Beijing, China
    \textsuperscript{\rm 3}Institute for Artificial Intelligence, Peking University, Beijing, China
    
    \texttt{\{shengyangli25, lysong25, 2501210709, 2201110745\}@stu.pku.edu.cn,}\\
    \texttt{\{dongyiting, li.cong, qingnishen, yuzf12\}@pku.edu.cn}\\
    
}

\begin{document}

\maketitle

\begin{abstract}

Federated learning enables collaborative model training across distributed edge devices while strictly preserving data privacy. To facilitate practical deployment on resource-constrained edge devices, Spiking Neural Networks (SNNs) have emerged as a promising alternative to traditional Artificial Neural Networks (ANNs) due to their sparse computing mechanisms and high energy efficiency. However, jointly training ANNs and SNNs exposes a challenge of representational misalignment, which is intrinsically caused by differences in information representation, specifically the semantic gap between continuous real-valued activations in ANNs and discrete spatio-temporal spikes in SNNs.
To overcome this barrier, we propose AS-FedBridge, a novel federated learning framework tailored for mixed ANN-SNN clients. AS-FedBridge features a lightweight Bridge equipped with a Pseudo-Spike Interface, which effectively projects continuous signals into a spike-compatible space to facilitate ANN-SNN alignment.
Given the absence of existing mixed ANN-SNN federated frameworks, we establish a comprehensive benchmark to evaluate against multiple advanced heterogeneous FL methods.
Our empirical analysis demonstrates a positive correlation between the degree of ANN-SNN alignment and the collaborative FL performance. Across four datasets, AS-FedBridge consistently demonstrates advanced accuracy while mitigating extreme scale, architecture, and client heterogeneity challenge.
Furthermore, our framework enables a highly controllable trade-off between model performance and resource efficiency. AS-FedBridge accomplishes these robust performance gains while introducing only marginal computational overhead, establishing a robust and practical foundation for mixed ANN-SNN federated learning systems.

\end{abstract}

\section{Introduction}

Federated Learning (FL) enables decentralized clients to collaboratively optimize a shared global objective without exposing their raw data \cite{mcmahan2017communication,kairouz2021advances}. While well-suited for data distributed across edge servers, mobile devices, and sensors, practical FL deployments face substantial heterogeneity in energy budgets and computational capacities  \cite{li2020federated,tan2022fedproto,li2019fedmd}. Typically, resource-rich clients can afford accurate but computationally costly models, whereas resource-constrained clients must rely on compact alternatives \cite{li2020federated}. Consequently, robust FL systems need to flexibly accommodate this heterogeneity to strike an optimal balance between global efficiency and accuracy.

\begin{figure}[tbp] 
\centering
\includegraphics[width=\columnwidth]{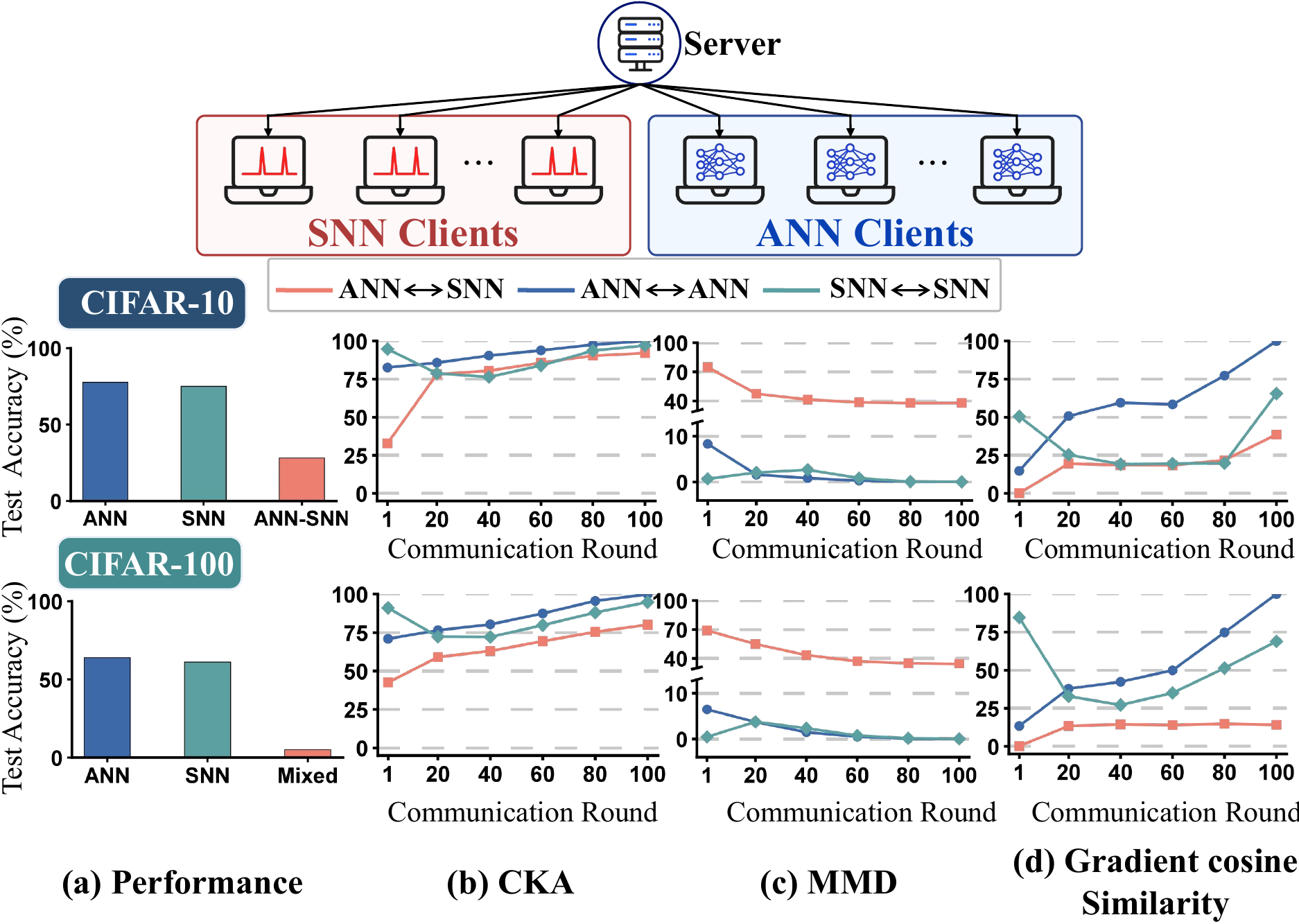}
\caption{Representation analysis of a mixed ANN-SNN federation. The severe accuracy drop is driven by fundamental mismatches in (b) Feature geometry (CKA), (c) Statistical distribution (MMD), (d) Optimization directions similarity. }
\label{fig:motivation}
\end{figure}

To accommodate resource-constrained clients, deploying quantized or binarized Artificial Neural Networks (ANNs) has become a standard paradigm \cite{courbariaux2014training,rastegari2016xnor,jacob2018quantization}. However, this approach suffers from a severe accuracy-efficiency trade-off. Aggressive compression techniques (low-bit quantization or heavy pruning) bottleneck the model's representational capacity and incur information loss \cite{dhar2021survey,sze2017efficient}, 
driving the need for fundamentally different low-power computing paradigms.
Instead of compromised ANNs, Spiking Neural Networks (SNNs) offer exceptional energy efficiency by communicating via binary spikes and exploiting sparse accumulation on neuromorphic hardware  \cite{fang2021deep,sengupta2019going,davies2018loihi}. Moreover, SNNs exploit temporal dynamics as an extra dimension to expand representational space, accumulating spatial features to enhance performance without inflating parameter counts \cite{fang2021deep,roy2019towards}. Coupled with recent training advances narrowing the accuracy gap on visual tasks \cite{fang2021deep,zheng2021going,sengupta2019going}, SNNs have emerged as ideal models for low-power clients. By retaining ANNs on resource-rich clients for high accuracy and fast convergence, a mixed federation containing both ANN and SNN clients can exploit complementary device capabilities, empowering heterogeneous clients to collaboratively learn from isolated data.

Existing heterogeneous FL methods implicitly assume that all clients operate within a compatible, continuous representation space. While exchanging intermediate features, prototypes, or logits successfully bypasses parameter aggregation \cite{mcmahan2017communication,li2020federated,wang2020tackling,hinton2015distilling,li2019fedmd,tan2022fedproto}, these techniques rely on a shared continuous semantic geometry, and \texttt{This assumption completely breaks down in mixed ANN-SNN federations}. While ANNs extract continuous, spatially organized features, SNNs encode information through discrete, temporally governed spikes\cite{fang2021deep}. Directly aligning these divergent signals induces severe representational mismatch and optimization conflicts.
We empirically validate this misalignment in Figure~\ref{fig:motivation} using  sampled non-IID subsets of CIFAR-10/100. Our analysis reveals a lack of topological similarity via Centered Kernel Alignment (CKA) \cite{kornblith2019similarity}, a massive distribution distance via Maximum Mean Discrepancy (MMD), and optimization conflicts evidenced by degraded Gradient Cosine Similarity. Consequently, naive FL triggers poor transfer, decimating the performance of mixed federations (Figure~\ref{fig:motivation}a). Therefore, a mixed federation must extract transferable knowledge while preserving continuous and spike-based local representations, motivating a shared intermediate mechanism for knowledge exchange.

To address this challenge, we propose \textbf{Federated Bridge Distillation for ANN-SNN Heterogeneity (AS-FedBridge)}. Rather than employing a conventional messenger model \cite{shen2020federated}, AS-FedBridge utilizes a lightweight Pseudo-Spike Regularized Bridge as a specialized mediator to accommodates between continuous ANN representations and sparse SNN features.   
Central to this transformation lies in the proposed Pseudo-Spike Interface, equipped with Pseudo-Spike Regularized (PSPR), which forces the pseudo-spike exchange activations to align with sparse SNN features. PSPR reshapes the Bridge into a unified, spike-compatible semantic space that preserves differentiable ANN optimization.  
We extensively evaluate AS-FedBridge under diverse federated configurations, encompassing Scale/Arch heterogeneity, and clients imbalance with advanced heterogeneous baselines. Empirical analysis (CKA/Gradient) establishes a positive correlation between the ANN-SNN distribution alignment and accuracy. Experiments confirm that actively bridging the ANN-SNN gap substantially improves representation similarity, mitigates optimization conflicts and achieving accuracy gains for ANN-SNN heterogeneous FL.


Our main contributions are summarized as follows:
\begin{itemize}
    \item We formulate the mixed ANN-SNN heterogeneous FL problems and establish a comprehensive evaluation benchmark, exploring the collaborative optimization under diverse non-IID and structural configurations.
    
    \item We propose AS-FedBridge, a novel ANN-SNN heterogeneous FL framework utilizing PSPR to align ANN-SNN features, enabling bidirectional knowledge exchange.
    
    \item Experiments demonstrate that AS-FedBridge mitigates representation misalignment between ANN and SNN, resolving optimization conflicts and achieving superior accuracy against advanced baselines.
\end{itemize}


\begin{figure*}[!t]
\centering
\includegraphics[width=0.85\textwidth]{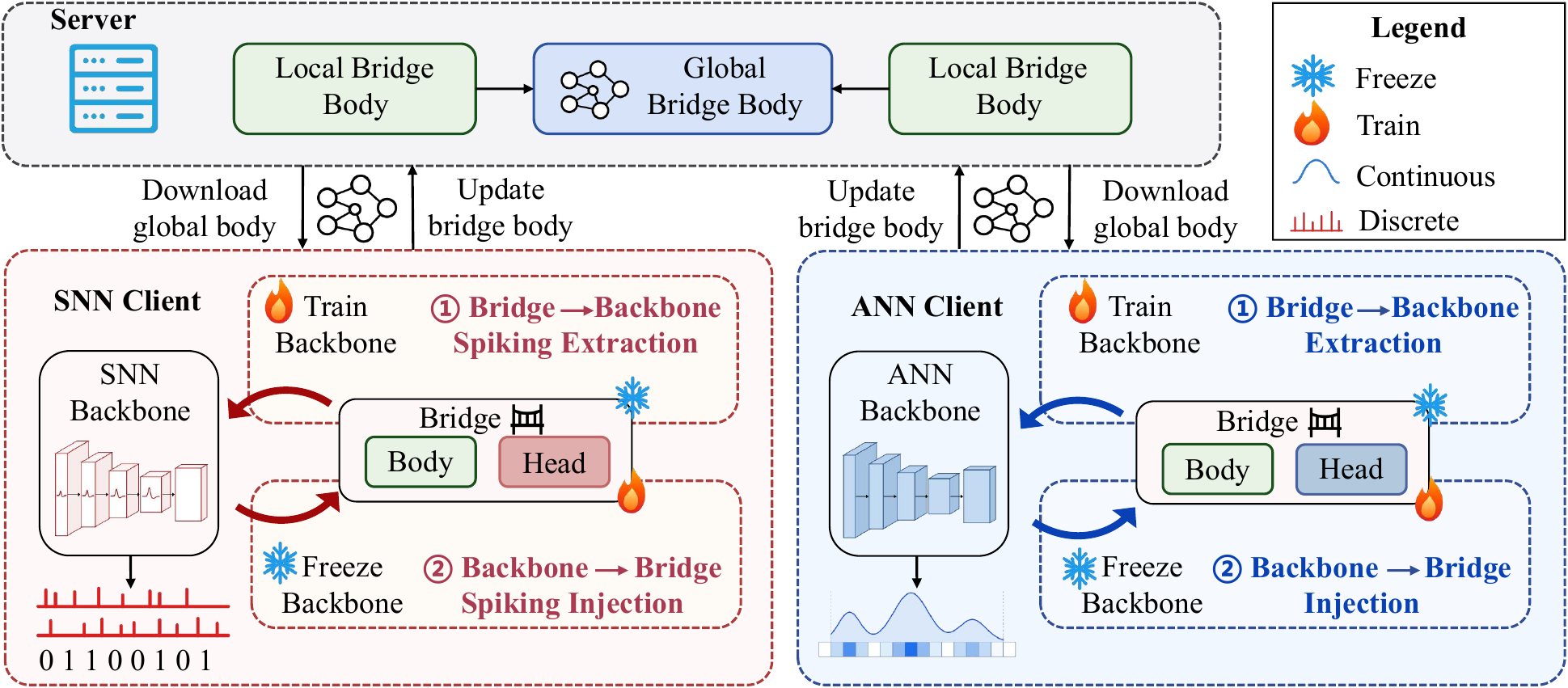}
\caption{Overall pipeline of AS-FedBridge. The server aggregates only the shared Bridge body, while heterogeneous ANN and SNN backbones remain local. In each round, clients first extract knowledge from the frozen global Bridge and then inject updated local knowledge into a local Bridge copy for aggregation.}
\label{fig:pipeline}
\end{figure*}

\section{Preliminary and Related Work}

\subsection{Heterogeneous Federated Learning}
Model-heterogeneous FL enables clients with different architectures or capacities to collaborate without directly aggregating incompatible backbones. Existing approaches exchange predictions through distillation \cite{li2019fedmd,shen2020federated}, communicate class-level prototypes \cite{tan2022fedproto,zhang2024fedtgp}, or coordinate through capacity-aware submodels and lightweight messengers \cite{kim2022depthfl,du2026fedfree,li2026feature}. However, these methods typically exchange continuous ANN-based logits, features, or prototypes and thus assume representation compatibility, leaving the continuous--spiking gap in mixed ANN-SNN federations unresolved.

\subsection{Federated Spiking Neural Networks}

Federated SNNs extend FL to spiking clients while accounting for temporal dynamics in distributed optimization. Early work applied FedAvg to homogeneous SNNs \cite{venkatesha2021federated}, followed by hierarchical communication and task-specific learning \cite{aouedi2023hfedsnn,xie2022efficient}, while recent methods address intra-SNN heterogeneity through label-skew-aware optimization and fusion across firing behaviors \cite{yu2024exploiting,tao2026sfedhifi}. However, these methods assume spiking or representation-compatible participants and cannot directly exchange knowledge with ANN clients, whose continuous outputs cannot be directly interpreted as temporal spike signals.

\subsection{ANN-SNN Alignment and Interaction}
Existing approaches align ANNs and SNNs through conversion from ANN activations to SNN firing rates using activation design, threshold calibration, membrane-potential correction, and error compensation \cite{ding2021optimal,bu2023optimal,hao2023reducing,jiang2023unified}, through hybrid architectures that co-locate both modalities \cite{aydin2024hybrid}, or through distillation from a fixed ANN teacher to an SNN student \cite{xu2023constructing,yu2025temporal,liu2026closer}. However, these settings target converted models, co-located pipelines, or fixed teacher--student pairs rather than collaboration among independently trained private models, while mismatches in rate coding, temporal dynamics, and feature distributions make direct feature or logit matching unreliable. AS-FedBridge instead learns a shared PSPR-regularized exchange space for bidirectional knowledge transfer across private ANN and SNN clients, without conversion, co-location, or a fixed teacher.

\section{Method}
\subsection{Problem Formulation and Bridge}
We consider $N$ clients with private datasets $\mathcal{D}_i$ and
backbones $f_i(x;\theta_i)$, where the client data may follow either
IID or non-IID distributions. Clients employ either continuous ANNs or
temporal SNNs; an SNN produces time-step logits
$\{Z_i^t\}_{t=1}^{T}$ with prediction
$\bar{Z}_i=T^{-1}\sum_t Z_i^t$. ANN and SNN backbones differ fundamentally in parameterization and information encoding, preventing joint aggregation and reliable alignment between continuous activations and sparse temporal spikes.

As illustrated in Figure~\ref{fig:pipeline}, we introduce a
lightweight Pseudo-spike Bridge $B(x;\phi_i)$ as the communication model to for knowledge exchange between ANN and SNN. Its parameters
$\phi_i=\{\omega,\psi_i\}$ consist of a globally aggregated body
$\omega$ and a private client head $\psi_i$. Each client first extracts
global knowledge from the Bridge into its private backbone and then
injects the updated local knowledge back into the Bridge. The Bridge
remains continuous for semantic transfer, while SNN clients
additionally align selected pseudo-spike activations with their
time-averaged firing rates, enabling the shared encoder to acquire
spike-compatible knowledge without being converted into an SNN.


\begin{figure}[t]
\centering
\includegraphics[width=\columnwidth]{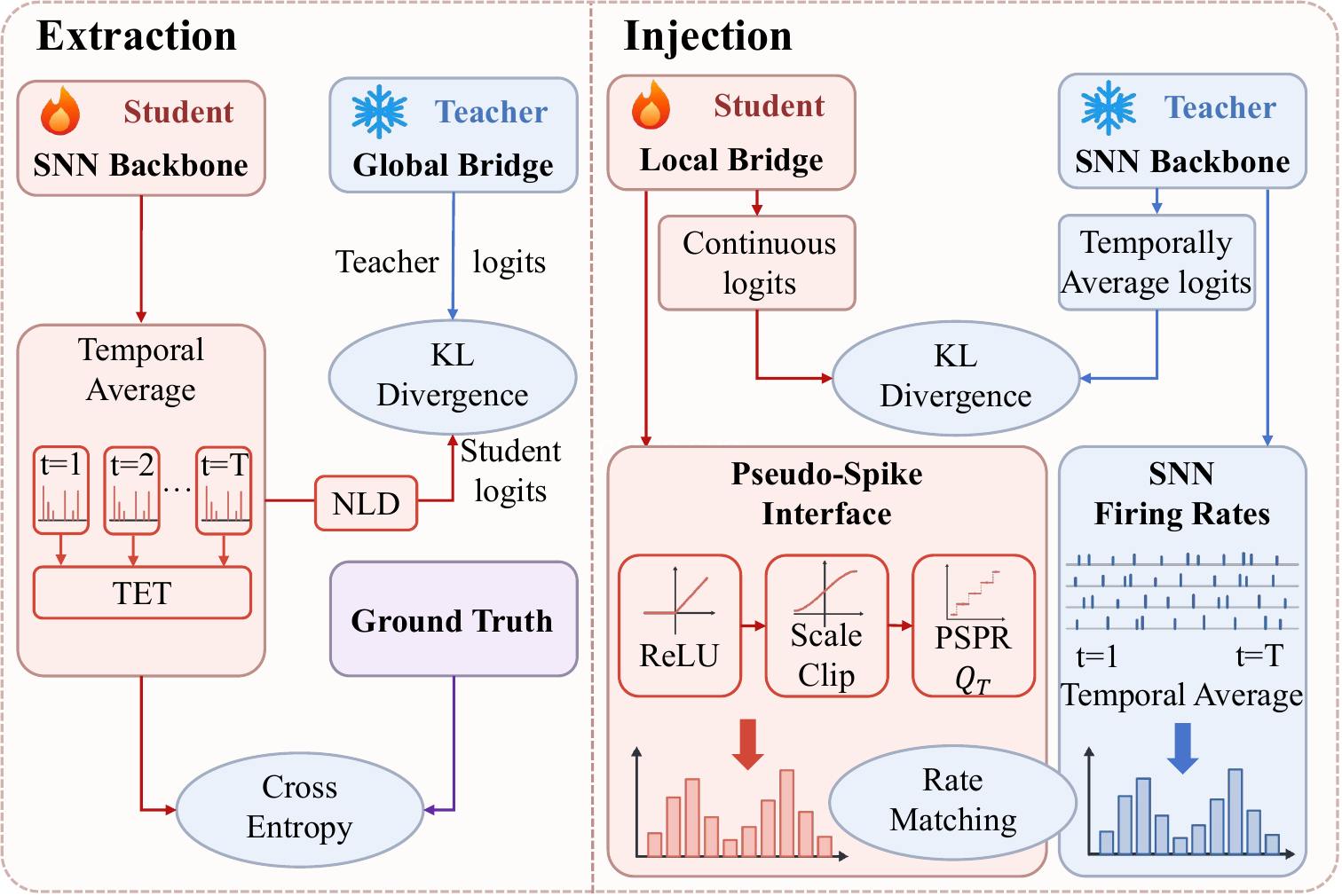}
\caption{SNN clients side bidirectional distillation in AS-FedBridge, including Bridge-to-SNN extraction and SNN-to-Bridge injection with bottleneck-level firing-rate matching and Pseudo-Spike Polarization Regularization (PSPR).}
\label{fig:extractors}
\end{figure}

\subsection{Pseudo-spike Interface}

The pseudo-spike interface transforms selected continuous Bridge
activations into spike-rate representations for SNN injection.
For each selected position $\ell \in \mathcal{P}$, we construct
a bounded rate variable
\begin{equation}
r_B^\ell
=
\left[
\frac{\mathrm{ReLU}(a_B^\ell)}{s_\ell}
\right]_{0}^{1},
\qquad
\ell \in \mathcal{P},
\label{eq:rate_port}
\end{equation}
where $s_\ell>0$ is a learnable channel-wise scaling factor and
$[\cdot]_{0}^{1}$ clips its input to $[0,1]$. The resulting
pseudo-spike path supports rate alignment, while the continuous
activations remain the Bridge classification path.

For an SNN operating over $T$ time steps, its empirical firing
rates lie in the discrete set
\begin{equation}
\mathcal{R}_T
=
\left\{
\frac{k}{T}
\;\middle|\;
k=0,\ldots,T
\right\}.
\label{eq:rate_set}
\end{equation}
We map each bounded Bridge rate onto this support through
\begin{equation}
\begin{aligned}
\tilde{r}_B^\ell
&=
r_B^\ell
+
\mathrm{sg}\!\left(
Q_T(r_B^\ell)-r_B^\ell
\right),\\
Q_T(r)
&=
\frac{1}{T}\,
\mathrm{round}(Tr).
\end{aligned}
\label{eq:rate_quantization}
\end{equation}
The stop-gradient construction produces discrete forward values
while retaining gradients to $r_B^\ell$ during backpropagation.

During SNN injection, the updated SNN is frozen, and its final
spike feature is projected into the same $d$-dimensional interface
as the designated Bridge bottleneck $\ell^\star \in \mathcal{P}$.
Let $S_i^t \in \{0,1\}^{d}$ denote this projected feature at time
$t$, with temporal firing rate
\begin{equation}
r_i
=
\frac{1}{T}
\sum_{t=1}^{T} S_i^t.
\label{eq:snn_rate}
\end{equation}
Using the frozen SNN rate as the target, we optimize
\begin{equation}
\mathcal{L}_{\mathrm{rate}}
=
\mathrm{MSE}\left(
\tilde{r}_B^{\ell^\star},
\mathrm{sg}\!\left(r_i\right)
\right).
\label{eq:rate_align}
\end{equation}
This shared interface supports firing-rate alignment across
heterogeneous backbones, while logit distillation transfers
class-level semantics. In the homogeneous setting,
Figure~\ref{fig:pseudospike_distribution} visualizes the Bridge
and SNN rate distributions at corresponding stages to analyze
their variation across network depth.

We further regularize the pre-quantized variables by controlling
their mean and variance:
\begin{equation}
\begin{aligned}
\mathcal{L}_{\mathrm{PSPR}}
={}&
\frac{1}{|\mathcal{P}|}
\sum_{\ell \in \mathcal{P}}
\mathbb{E}\Bigl[
\bigl(
|\mu_\ell-V_{\mathrm{th}}|
-k_d\sigma_\ell
\bigr)_{+}^{2}
\\
&\qquad\qquad+
\lambda_{\mathrm{var}}
\bigl(
\sigma_{\min}-\sigma_\ell
\bigr)_{+}^{2}
\Bigr].
\end{aligned}
\label{eq:pspr}
\end{equation}
Here, $(u)_{+}=\max(u,0)$, while $\mu_\ell$ and $\sigma_\ell$
are the channel-wise mean and standard deviation of $r_B^\ell$.
The first term encourages $V_{\mathrm{th}}$ to lie within
$k_d\sigma_\ell$ of $\mu_\ell$, while the second encourages
$\sigma_\ell$ to remain above $\sigma_{\min}$. PSPR encourages
sufficient variation before $Q_T$ maps the variables onto
$\mathcal{R}_T$, and $\mathcal{L}_{\mathrm{rate}}$ aligns the
bottleneck rates with the observed SNN activity.

\subsection{Bidirectional Bridge Distillation}

Figure~\ref{fig:extractors} illustrates the extraction and injection
paths for an SNN client. At round $r$, each client combines the
broadcast Bridge body $\omega^r$ with its private head $\psi_i$.
The frozen Bridge first supervises the private backbone during
extraction; the updated backbone then serves as the frozen teacher
of the local Bridge during injection.

For student logits $Z_s$ and teacher logits $Z_t$, we define

\begin{align}
\mathcal{L}_{\mathrm{KD}}(Z_s,Z_t)
={}&
\tau^2 D_{\mathrm{KL}}\Bigl(
\mathrm{softmax}(Z_t/\tau)
\notag\\[-2pt]
&\qquad
\mathrel{\|}
\mathrm{softmax}(Z_s/\tau)
\Bigr).
\label{eq:kd}
\end{align}

where $\tau$ is the distillation temperature.

\textbf{Extraction.}
For an ANN client, the frozen Bridge transfers global class semantics
through standard supervised distillation:
\begin{equation}
\mathcal{L}_{Ext}^{ann}
=
\mathcal{L}_{CE}(Z_i,y)
+
\alpha_{kd}\mathcal{L}_{KD}(Z_i,Z_B),
\label{eq:ann_extraction}
\end{equation}
where $Z_i$ and $Z_B$ are the ANN and Bridge logits, respectively.

For an SNN client, we combine \emph{Temporal Efficient Training}
(TET)~\cite{deng2022temporal} with a centered variant of
\emph{Noise-smoothed Logits Distillation}
(NLD)~\cite{liu2026closer}. Given the time-step logits
$\{Z_i^t\}_{t=1}^{T}$, let
$\bar Z_i=T^{-1}\sum_{t=1}^{T}Z_i^t$ and perturb only their temporal
average:
\begin{equation}
\begin{aligned}
\epsilon_i&\sim\mathcal{N}(0,I),\\
\hat Z_i&=
\bar Z_i+
\lambda_{nld}\sg(\sigma_i)\epsilon_i,
\end{aligned}
\label{eq:nld}
\end{equation}
where $\sigma_i$ is the per-sample standard deviation of $\bar Z_i$
over classes, and $\sg(\cdot)$ blocks gradients through the noise
scale. The SNN extraction objective is
\begin{equation}
\begin{aligned}
\mathcal{L}_{Ext}^{snn}
={}&
\frac{1}{T}\sum_{t=1}^{T}
\Big[
(1-\lambda_{tet})
\mathcal{L}_{CE}(Z_i^t,y)\\
&\qquad+
\lambda_{tet}
\MSE(Z_i^t,V_{tet})
\Big]\\
&+
\alpha_{kd}
\mathcal{L}_{KD}(\hat Z_i,Z_B),
\end{aligned}
\label{eq:snn_extraction}
\end{equation}
where $V_{tet}$ is the reference output used by TET. TET supervises
the original time-step logits, whereas only the NLD-perturbed average
is used for distillation.

\textbf{Injection.}
After extraction, the updated private backbone is frozen and becomes
the teacher of the Bridge. For an ANN client, injection transfers
continuous class semantics through
\begin{equation}
\begin{aligned}
\mathcal{L}_{Inj}^{ann}
={}&
\alpha_{teach}
\mathcal{L}_{KD}(Z_B,Z_i)
+
\alpha_{ce}
\mathcal{L}_{CE}(Z_B,y)\\
&+
\alpha_{prox}
\|\omega_i-\omega^r\|_2^2.
\end{aligned}
\label{eq:ann_injection}
\end{equation}

For an SNN client, the temporally averaged logits provide semantic
supervision, while the averaged hidden spikes calibrate the
pseudo-spike interface:
\begin{equation}
\begin{aligned}
\mathcal{L}_{Inj}^{snn}
={}&
\alpha_{teach}
\mathcal{L}_{KD}(Z_B,\bar Z_i)
+
\alpha_{ce}
\mathcal{L}_{CE}(Z_B,y)\\
&+
\alpha_{rate}
\mathcal{L}_{rate}
+
\alpha_{pspr}
\mathcal{L}_{PSPR}\\
&+
\alpha_{prox}
\|\omega_i-\omega^r\|_2^2.
\end{aligned}
\label{eq:snn_injection}
\end{equation}
The two SNN-specific terms respectively align firing rates and
regularize the pseudo-spike distribution.

After injection, only the Bridge body is uploaded and aggregated:
\begin{equation}
\omega^{r+1}
=
\sum_{i\in\mathcal{S}_r}
\frac{|\mathcal{D}_i|}
{\sum_{j\in\mathcal{S}_r}|\mathcal{D}_j|}
\omega_i,
\label{eq:aggregation}
\end{equation}
where $\mathcal{S}_r$ denotes the participating clients; all private
backbones and Bridge heads remain local.

\section{Experiments}

\begin{figure*}[!t]
\centering
\includegraphics[width=0.95\textwidth]{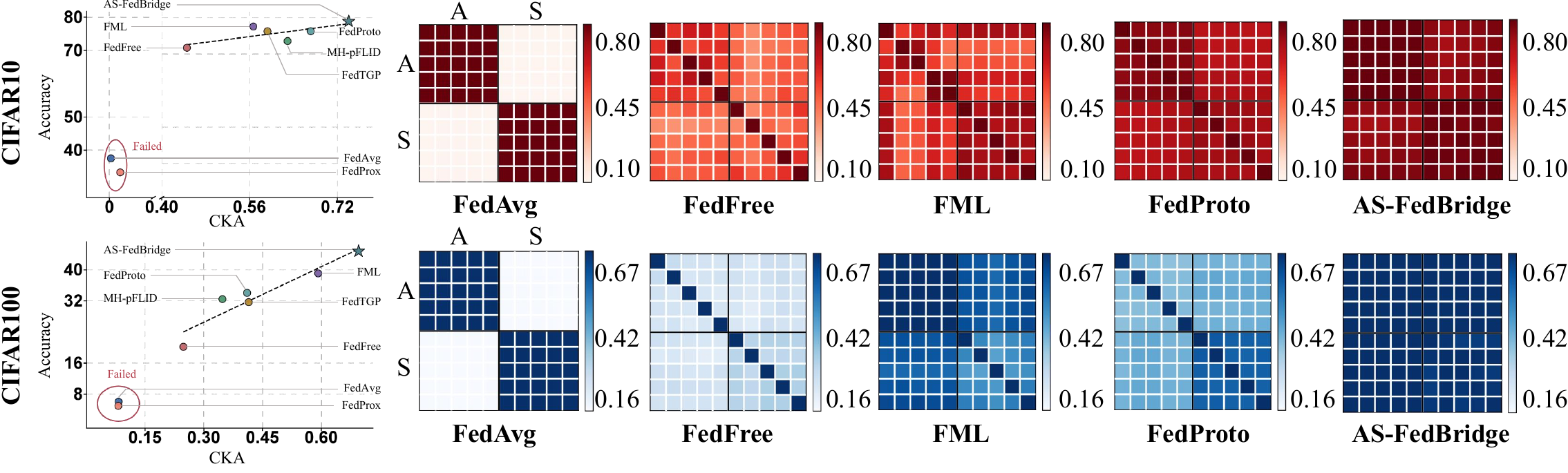}
\caption{Representation alignment evaluated via cross-client CKA alongside model accuracy on CIFAR-10 and CIFAR-100. AS-FedBridge substantially strengthens both intra-group and cross-group similarity to drive collaborative accuracy.}
\label{fig:cka}

\centering
\includegraphics[width=0.95\textwidth]{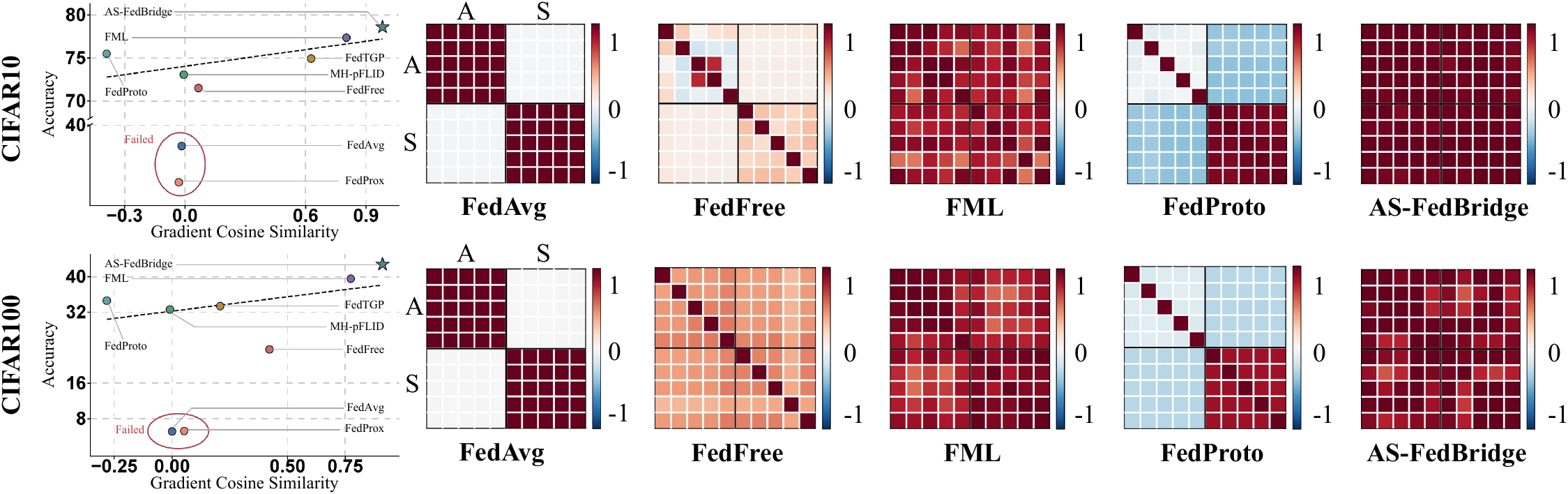}
\caption{Gradient cosine similarity alongside model accuracy on CIFAR-10 and CIFAR-100. AS-FedBridge ensures broadly positive gradient alignment across all ANN and SNN clients to prevent conflicting parameter updates.}
\label{fig:gradient_consistency}
\end{figure*}

\begin{figure}[!t]
\centering
\includegraphics[width=0.95\columnwidth]{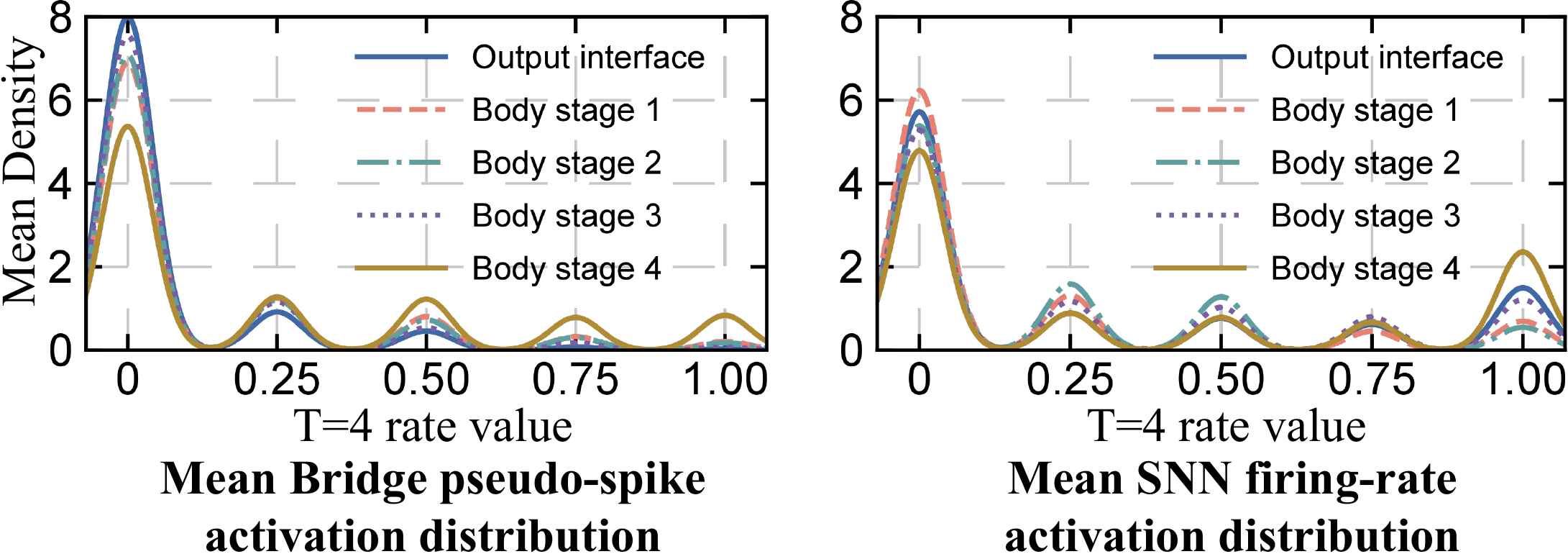}
\caption{Layer-wise distributions of Bridge pseudo-spike activations
and SNN firing rates for a non-IID client ($T=4$).}
\label{fig:pseudospike_distribution}
\end{figure}

\subsection{Experimental Setup}
\textbf{Datasets and Partition.}
We conduct extensive evaluations across three static vision benchmarks (CIFAR-10, CIFAR-100, Tiny-ImageNet) and one neuromorphic dataset (CIFAR10-DVS) \cite{krizhevsky2009learning,deng2009imagenet,li2017cifar10}. 
Unless stated otherwise, ten clients follow a Dirichlet partition with $\alpha=0.1$, and each client retains a private local test split. More ablations of $\alpha$ are in Appendix.

\noindent\textbf{Models.}
The default federated environment simulates a balanced mixed ANN-SNN scenario comprising five continuous ResNet-18 clients and five Spiking-ResNet18 clients operating at $T=4$ time steps. To comprehensively measure collaborative performance, we report the mean top-1 accuracy independently for the ANN group, the SNN group, and the overall federation.

\noindent\textbf{Baselines.}
Because no prior benchmark exists for mixed ANN-SNN FL, we establish a comprehensive ANN-SNN evaluation benchmark, which defines performance boundaries using Centralized and Standalone alongside isolated FedAvg and FedProx. Furthermore, We reproduce six advanced heterogeneous FL methods (FedFree, FedProto, FedTGP, FML, MH-pFLID, SFedHIFI) \cite{shen2020federated,tan2022fedproto,zhang2024fedtgp,Xie24,du2026fedfree,tao2026sfedhifi} to serve as advanced mixed ANN-SNN baselines.

\subsection{Mechanism Analysis}
\begin{table*}[t]
\centering
{\small
\setlength{\tabcolsep}{1.8pt}
\renewcommand{\arraystretch}{1.08}
\begin{tabular*}{\textwidth}{
@{\extracolsep{\fill}}ll*{4}{ccc}@{}
}
\toprule
\multirow{2}{*}{\textbf{Paradigm}} &
\multirow{2}{*}{\textbf{Method}} &
\multicolumn{3}{c}{\textbf{CIFAR-10}} &
\multicolumn{3}{c}{\textbf{CIFAR-100}} &
\multicolumn{3}{c}{\textbf{Tiny-ImageNet}} &
\multicolumn{3}{c}{\textbf{CIFAR10-DVS}} \\
\cmidrule(lr){3-5}\cmidrule(lr){6-8}
\cmidrule(lr){9-11}\cmidrule(lr){12-14}
& & ANN & SNN & Avg. & ANN & SNN & Avg. &
ANN & SNN & Avg. & ANN & SNN & Avg. \\
\midrule
\multirow{2}{*}{Non-FL}
& Centralized
& 95.05 & 93.37 & 94.21
& 78.73 & 74.66 & 76.70
& 60.90 & 57.21 & 59.06
& 82.40 & 72.80 & 77.60 \\
& Standalone
& 88.66 & 84.82 & 86.74
& 62.79 & 56.61 & 59.70
& 56.60 & 46.43 & 51.52
& 76.66 & 78.88 & 77.77 \\
\midrule
\multirow{2}{*}{\shortstack[l]{Isolated}}
& Isolated FedAvg
& 78.13 & 83.21 & 80.67
& 63.80 & 58.44 & 61.12
& 49.66 & 48.02 & 48.84
& 51.51 & 54.92 & 53.90 \\
& Isolated FedProx
& 76.16 & 82.69 & 79.43
& 64.42 & 55.86 & 60.14
& 49.90 & 40.70 & 45.30
& 50.17 & 52.64 & 51.90 \\
\midrule
\multirow{3}{*}{Global-Model}
& FedAvg
& 40.20 & 16.04 & 28.12
& 6.41 & 3.73 & 5.07
& 1.42 & 0.76 & 1.09
& 5.69 & 12.98 & 10.80 \\
& FedProx
& 37.75 & 23.57 & 30.66
& 5.67 & 3.53 & 4.60
& 1.78 & 1.44 & 1.61
& 27.76 & 12.27 & 16.90 \\
& FedFree
& 88.95 & 60.58 & 74.77
& 61.64 & 54.18 & 57.91
& 44.11 & 43.15 & 45.08
& 29.42 & 46.16 & 37.79 \\
\midrule
\multirow{2}{*}{Prototype}
& FedProto
& 90.16 & 89.85 & 90.01
& 65.53 & 71.02 & 68.28
& 58.64 & 48.63 & 53.63
& 82.03 & 78.13 & \underline{80.08} \\
& FedTGP
& 90.97 & 88.11 & 89.54
& 68.43 & 63.57 & 66.00
& 58.51 & 49.47 & 53.99
& 77.50 & 70.32 & 72.40 \\
\midrule
\multirow{3}{*}{Shared-Model}
& FML
& 91.70 & 89.50 & \underline{90.60}
& 69.42 & 68.08 & \underline{68.75}
& 57.98 & 53.34 & \underline{55.66}
& 82.37 & 74.43 & 78.40 \\
& MH-pFLID
& 90.69 & 87.09 & 88.89
& 66.29 & 63.99 & 65.14
& 58.27 & 50.98 & 54.63
& 71.97 & 75.19 & 73.58 \\
\rowcolor{gray!20}[6pt][6pt]
\cellcolor{white}
& \textbf{AS-FedBridge}
& 91.53 & 90.54 & \textbf{91.12}
& 72.21 & 71.08 & \textbf{71.64}
& 55.96 & 56.27 & \textbf{56.12}
& 83.15 & 80.98 & \textbf{82.06} \\
\bottomrule
\end{tabular*}
}
\caption{Performance evaluation of fixed ANN and SNN clients under a Dirichlet non-IID partitioning at $\alpha=0.1$. ``Avg.'' averages the participating clients. Bold and underline values highlight the best and second-best results respectively.}
\label{tab:main_results}
\end{table*}

\begin{table*}[t]
\centering
{\small
\setlength{\tabcolsep}{2.0pt}
\renewcommand{\arraystretch}{1.08}
\begin{tabular*}{\textwidth}{
@{\extracolsep{\fill}}ll*{4}{ccc}@{}
}
\toprule
\multirow{3}{*}{\textbf{Paradigm}} &
\multirow{3}{*}{\textbf{Method}} &
\multicolumn{6}{c}{\textbf{(A) Scale Heterogeneity}} &
\multicolumn{6}{c}{\textbf{(B) Architecture Heterogeneity}} \\
\cmidrule(lr){3-8}\cmidrule(lr){9-14}
& & \multicolumn{3}{c}{\textbf{CIFAR-10}} &
\multicolumn{3}{c}{\textbf{CIFAR-100}} &
\multicolumn{3}{c}{\textbf{CIFAR-10}} &
\multicolumn{3}{c}{\textbf{CIFAR-100}} \\
\cmidrule(lr){3-5}\cmidrule(lr){6-8}
\cmidrule(lr){9-11}\cmidrule(lr){12-14}
& & ANN & SNN & Avg. & ANN & SNN & Avg. &
ANN & SNN & Avg. & ANN & SNN & Avg. \\
\midrule
Non-FL & Standalone
& 88.67 & 82.41 & 85.54
& 60.87 & 49.78 & 55.33
& 90.09 & 86.03 & 88.06
& 65.31 & 61.44 & 63.38 \\
\midrule
Isolated & SFedHIFI$^{*}$
& 89.82 & 69.56 & 79.69
& 63.93 & 56.33 & 60.13
& - & - & -
& - & - & - \\
\midrule
Global-Model & FedFree
& 89.17 & 69.47 & 79.32
& 59.37 & 54.91 & 57.14
& 67.10 & 69.12 & 68.11
& 61.43 & 61.71 & 61.57 \\
\midrule
\multirow{2}{*}{Prototype}
& FedProto
& 89.05 & 89.29 & \underline{89.17}
& 65.68 & 62.80 & 64.24
& 89.56 & 89.88 & \underline{89.72}
& 66.68 & 68.79 & 67.74 \\
& FedTGP
& 91.80 & 84.36 & 88.08
& 63.42 & 61.73 & 62.58
& 68.79 & 85.84 & 77.32
& 66.38 & 63.72 & 65.05 \\
\midrule
\multirow{3}{*}{Shared-Model}
& FML
& 90.38 & 85.91 & 88.14
& 67.34 & 66.54 & \underline{66.94}
& 91.24 & 87.56 & 89.40
& 69.83 & 66.77 & \underline{68.30} \\
& MH-pFLID
& 89.22 & 81.24 & 85.23
& 62.99 & 61.79 & 62.39
& 91.43 & 87.04 & 89.24
& 65.85 & 63.35 & 64.60 \\
\rowcolor{gray!20}[6pt][6pt]
\cellcolor{white}
& \textbf{AS-FedBridge}
& 91.38 & 88.97 & \textbf{90.18}
& 73.24 & 72.77 & \textbf{73.00}
& 92.45 & 91.57 & \textbf{91.88}
& 73.59 & 72.62 & \textbf{72.79} \\
\bottomrule
\end{tabular*}
}
\caption{Performance evaluation under scale and architecture heterogeneity with Dirichlet $\alpha=0.1$. ($*$) indicates spiking FL method re-implemented under our ANN-SNN experimental setup. Bold and underline indicate best and second-best results.}
\label{tab:heterogeneous}
\end{table*}

\noindent\textbf{Representation Alignment vs. Accuracy.}
We utilize Centered Kernel Alignment (CKA) \cite{kornblith2019similarity} to correlate representation similarity with model accuracy (Figure~\ref{fig:cka}). Experiments show that Direct aggregation preserves intra-group similarity but leaves ANN and SNN clients in severely disjoint representation spaces, which causes performance collapse. AS-FedBridge successfully bridges this semantic barrier, yielding densely aligned CKA blocks both within and across distinct ANN-SNN group. This analysis establishes a positive correlation where higher ANN-SNN similarity more-likely yields higher global accuracy, proving the absolute necessity of our shared Bridge.


\noindent\textbf{Gradient Consistency Analysis. }
Figure~\ref{fig:gradient_consistency} evaluates gradient cosine similarity to link optimization consistency with final model performance. While baseline methods suffer from near-zero or negative gradients between ANN and SNN that trigger destructive interference, AS-FedBridge yields broadly positive similarities. By providing a unified feature compatible space, our framework prevents continuous ANNs and sparse SNNs updates from inducing gradient conflicts, resolving severe optimization conflicts.s, resolving severe optimization conflicts.

\noindent\textbf{Pseudo-spike Output Distribution. }
Figure~\ref{fig:pseudospike_distribution} contrasts the Bridge pseudo-spike outputs against time-averaged SNN firing rates across four body stages and the output interface. Both distributions exhibit prominent peaks at $\mathcal{R}_4=\{0, 0.25, 0.5, 0.75, 1\}$, while their probability masses shift dynamically across network layers. This structural variation demonstrates that the Bridge successfully captures layer-dependent temporal firing features rather than collapsing into a static multi-peak pattern, thereby supplying a robust rate-compatible representation for effective rate matching during SNN feature injection.

\subsection{Main Results with Mixed ANN and SNN}

\begin{table}[t]
\centering
{\small
\setlength{\tabcolsep}{1.2pt}
\renewcommand{\arraystretch}{1.10}
\begin{tabularx}{\columnwidth}{
@{}>{\raggedright\arraybackslash}Xlcrr@{}
}
\toprule
& & & \multicolumn{2}{c}{\textbf{Computation (M)}} \\
\cmidrule(lr){4-5}
\textbf{Setting} & \textbf{Method} & \textbf{Acc.} $\uparrow$ &
\textbf{FLOPs} $\downarrow$ & \textbf{SOPs} $\downarrow$ \\
\midrule
\multicolumn{5}{@{}l}{\textit{Homogeneous Collaboration}} \\
Binary-ANN & FedAvg & 41.25 & 3.85$^\dagger$ & 0 \\
SNN ($T=2$) & FedAvg & 59.39 & 0 & 81.69 \\
SNN ($T=4$) & FedAvg & 61.06 & 0 & 167.17 \\
SNN ($T=8$) & FedAvg & 61.81 & 0 & 338.29 \\
ANN & FedAvg & 63.98 & 1111.15 & 0 \\
\midrule
\multicolumn{5}{@{}l}{\textit{Mixed Collaboration}} \\
Binary--ANN & Bridge & 60.84 & 655.59$^\ddagger$ & 0 \\
\rowcolor{gray!20}
\textbf{ANN-SNN ($T=2$)} & \textbf{AS-FedBridge} & 67.01 & 653.67 & 40.85 \\
\rowcolor{gray!20}
\textbf{ANN-SNN ($T=4$)} & \textbf{AS-FedBridge} & 71.64 & 653.67 & 83.58 \\
\rowcolor{gray!20}
\textbf{ANN-SNN ($T=8$)} & \textbf{AS-FedBridge} & 72.49 & 653.67 & 169.15 \\
ANN--ANN & Bridge & 72.67 & 1209.24 & 0 \\
\bottomrule
\end{tabularx}
}
\caption{Performance and computation costs on CIFAR-100 $\alpha=0.1$.
 $^\dagger$ and $^\ddagger$ indicate an additional 553.65M and 276.82M XNOR-popcount operations respectively.}
\label{tab:mixed_necessity}
\end{table}

\begin{table}[t]
\centering
{\small
\setlength{\tabcolsep}{3.5pt}
\renewcommand{\arraystretch}{1.10}
\begin{tabularx}{\columnwidth}{
@{}>{\raggedright\arraybackslash}Xccc@{}
}
\toprule
\textbf{Ablation Variant} & \textbf{ANN} & \textbf{SNN} &
\textbf{Avg.} \\
\midrule
\multicolumn{4}{@{}l}{\textit{Unidirectional Transfer}} \\
ANN $\rightarrow$ Bridge $\rightarrow$ SNN only
& 65.36 & 70.87 & 68.12 \\
SNN $\rightarrow$ Bridge $\rightarrow$ ANN only
& 65.17 & 71.43 & 68.30 \\
\midrule
\multicolumn{4}{@{}l}{\textit{Component Ablation}} \\
w/o Pseudo-spike Interface
& 68.01 & 71.64 & \underline{69.83} \\
w/o Personalized Bridge Head
& 65.85 & 70.44 & 68.15 \\
\midrule
\rowcolor{gray!20}

\textbf{AS-FedBridge}
& 72.21  & 71.08 & \textbf{71.64} \\
\bottomrule
\end{tabularx}
}
\caption{Ablation study on CIFAR-100 with $\alpha=0.1$. }
\label{tab:ablation}
\end{table}

\begin{table}[t]
\centering
{\small
\setlength{\tabcolsep}{1.2pt}
\renewcommand{\arraystretch}{1.06}

\begin{tabularx}{\columnwidth}{llXrrr}
\toprule
&
&
\multicolumn{2}{c}{\textbf{Communication}}
&
\multicolumn{2}{c}{\textbf{Computation} $\downarrow$}
\\
\cmidrule(lr){3-4}
\cmidrule(l){5-6}

\textbf{Paradigm}
&
\textbf{Method}
&
\textbf{Content}
&
\textbf{MB} $\downarrow$
&
\textbf{ANN}
&
\textbf{SNN}
\\
\midrule

\multirow{3}{*}{%
  \shortstack[l]{\textit{Global}\\\textit{Model}}}
& FedAvg  & Full    & 43.20 & 16.67 & 3.09 \\
& FedProx & Full    & 43.20 & 16.67 & 5.77 \\
& FedFree & Top-$k$ & 18.00 & 16.67 & 2.91 \\

\addlinespace[2pt]
\multirow{2}{*}{\textit{Prototype}}
& FedProto & Class     & 0.10 & 17.78 & 3.09 \\
& FedTGP   & Trainable & 0.60 & 17.78 & 2.96 \\

\addlinespace[2pt]
& FML
& Distiller
& 3.93
& 18.14
& 3.06 + 1.47 \\

& MH-pFLID
& Messenger
& 3.93
& 18.57
& 3.01 + 0.79 \\

\rowcolor{gray!20}
\cellcolor{white}
\multirow{-3}{*}{%
  \shortstack[l]{\textit{Shared}\\\textit{Model}}}
& \textbf{AS-FedBridge}
& \textbf{Bridge}
& \textbf{3.93}
& \textbf{18.57}
& \textbf{3.21 + 0.79} \\
\bottomrule
\end{tabularx}
}
\caption{Per-client communication and computation costs under a 5:5
ANN-SNN split. Communication is reported in MB, ANN computation in
GFLOPs, and SNN computation in GSOPs + dense-model GFLOPs when
applicable.}
\label{tab:comm_cost}
\end{table}

\noindent\textbf{ANN–SNN Heterogeneity. }
We initially benchmark mixed collaboration using identical ResNet-18 structures for all ANN clients and Spiking-ResNet18 for all SNN clients to isolate the semantic divide. To ensure a fair comparison, we reproduced all baseline federated learning methods to specifically accommodate this ANN-SNN mixed heterogeneous spiking FL paradigm.
Table~\ref{tab:main_results} confirms that AS-FedBridge comprehensively outperforms all baselines with peak average accuracies of 91.12\%, 71.64\%, 56.12\%, and 82.06\% across the four datasets. AS-FedBridge exceeds the strongest alternative methods by up to 2.89 percentage points, whereas parameter aggregations like FedAvg suffer collapse directly caused by architectural misalignment.

\begin{figure*}[!t]
\centering
\includegraphics[width=\textwidth]{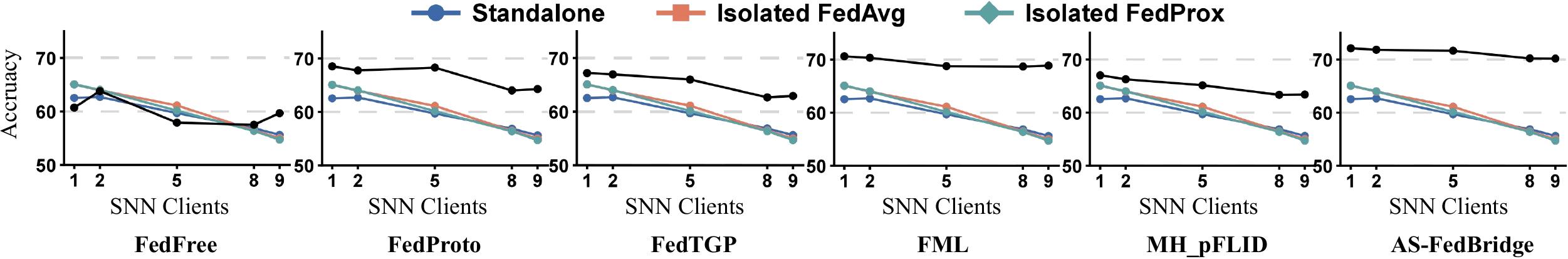}
\caption{Impact of varying SNN client numbers on overall accuracy within a ten-client CIFAR-100 federation.}
\label{fig:pareto_distribution}
\end{figure*}

\noindent\textbf{Scale and Architecture Heterogeneity. }
To demonstrate robustness beyond uniform backbones, we further evaluate our framework under severe structural variant by introducing scale heterogeneity with diverse architecture width multipliers  ($0.25\times$ to $1.0\times$ width multipliers) alongside architecture heterogeneity combining VGG, MobileNet, and ResNet. Table~\ref{tab:heterogeneous} confirms that AS-FedBridge dominates both complex scenarios by achieving the highest collaborative accuracy.


\noindent\textbf{Clients Imbalance Heterogeneity. }
Figure~\ref{fig:pareto_distribution} investigates collaborative stability by varying the number of participating SNN clients from one to nine. While baseline methods suffer severe accuracy degradation or unpredictable fluctuations as the SNN group increasingly dominates, AS-FedBridge maintains a highly stable trajectory near 70\% accuracy across all group ratios. This consistent superiority proves that our Pseudo-Spike Regularize Bridge effectively neutralizes structural bias and prevents performance collapse independently of the prevailing client distribution.

\subsection{Necessity of Mixed ANN-SNN Collaboration}
Table~\ref{tab:mixed_necessity} validates the necessity of mixed collaborative learning by contrasting our paradigm against homogeneous and low-bit alternatives. At $T=4$, AS-FedBridge achieves a remarkable 71.64\% accuracy, comprehensively eclipsing ANN-, SNN-only Homogeneous FL, and Binary--ANN frameworks by margins of 7.66, 10.58, and 10.80. This performance leap proves that ANN-SNN federations unlock positive knowledge transfer absent in isolated methods.  Furthermore, while an ANN--ANN Bridge yields higher accuracy, it incurs high floating-point computational overhead. Evaluating multiple timesteps demonstrates that modulating $T$ enables accuracy enhancements without expanding the parameter footprint\cite{sengupta2019going}. This temporal scaling grants AS-FedBridge a flexible performance-efficiency tradeoff that adapts to constrained edge deployments.

\subsection{Ablation Study}
Table~\ref{tab:ablation} confirms the essential contribution of every proposed architectural module. Constraining knowledge transfer to a unidirectional pathway reduces accuracy by at least 3.34, establishing that bidirectional extraction and injection are required for mixed ANN-SNN synergy. Removing the personalized Bridge head causes a 3.49 drop, highlighting its function in mitigating representational personalized divergence. Finally, excluding the Pseudo-spike Interface decreases accuracy by 1.81, confirming the efficiency of the firing-rate alignment mechanisms previously captured in Figure~\ref{fig:pseudospike_distribution}.

\subsection{Communication and Training Cost}
Table~\ref{tab:comm_cost} demonstrates that AS-FedBridge achieves exceptional mixed ANN-SNN alignment while bounding both communication and computational overhead. By retaining all private bodies and personalized heads on local devices, our framework transmits a highly compact 3.93 MB shared Bridge body. While prototype methods communicate smaller non-trainable summaries, our method provides a deeply trainable representation required for severe heterogeneity. 
Furthermore, processing this localized auxiliary bridge introduces minor additional computational overhead during the training phase and this shared mediator scales independently of the diverse private client architectures.


\section{Conclusion}
We propose AS-FedBridge to resolve the ANN-SNN representation mismatch in mixed ANN-SNN federated learning through a lightweight shared Bridge and a novel Pseudo-Spike Interface. Extensive experiments demonstrate that our framework achieves state-of-the-art accuracy and exceptional computational efficiency under severe heterogeneity.

\bibliography{aaai2027,reference}
\clearpage
\appendix
\setcounter{secnumdepth}{2}

\section{Additional Experiments}
\label{app:empirical}

This section presents five supplementary empirical studies. First, we add more component ablation experiments. Second, we evaluate
AS-FedBridge under IID and moderate non-IID partitions
($\alpha=0.5$) to assess its robustness to different degrees of data
heterogeneity. Third, we use squared Maximum Mean Discrepancy ($\operatorname{MMD}^{2}$) to
evaluate the feature-distribution gap between ANN and SNN
clients. Fourth, matched 2-, 4-, and 8-bit quantized-network controls
examine whether conventional low-precision models can replace SNN clients. Finally, we analyze the sensitivity of AS-FedBridge to key Bridge
configurations. Detailed experimental setup and theoretical clarifications are
provided in Sections \textbf{Experimental Details} and \textbf{Method Details}, respectively.

\subsection{Additional Component Ablations}
\label{app:ablation_results}

Table~\ref{tab:ablation} further decomposes the two transfer directions
and the Pseudo-spike Interface. All variants follow the same data
partition, model configuration, and training protocol, while retaining
local supervised learning. ANN $\rightarrow$ Bridge $\rightarrow$ SNN
keeps ANN injection and SNN extraction, whereas the reverse variant
keeps SNN injection and ANN extraction.

For the component ablations, w/o Pseudo-spike Interface removes the
entire rate branch, including PSPR, $Q_T$, and
$\mathcal{L}_{\mathrm{rate}}$, while preserving continuous logit
distillation. Removing the personalized Bridge head makes the classifier
globally aggregated. The remaining variants individually remove
distribution regularization, discrete rate projection, or explicit
bottleneck rate alignment.

\begin{table}[ht]
\centering
{\small
\setlength{\tabcolsep}{3.5pt}
\renewcommand{\arraystretch}{1.10}
\begin{tabularx}{\columnwidth}{
@{}>{\raggedright\arraybackslash}Xccc@{}
}
\toprule
\textbf{Ablation Variant} & \textbf{ANN} & \textbf{SNN} &
\textbf{Avg.} \\
\midrule
\multicolumn{4}{@{}l}{\textit{Unidirectional Transfer}} \\
ANN $\rightarrow$ Bridge $\rightarrow$ SNN only
& 65.36 & 70.87 & 68.12 \\
SNN $\rightarrow$ Bridge $\rightarrow$ ANN only
& 65.17 & 71.43 & 68.30 \\
\midrule
\multicolumn{4}{@{}l}{\textit{Component Ablation}} \\
w/o Pseudo-spike Interface
& 68.01 & 71.64 & 69.83 \\
w/o Personalized Bridge Head
& 65.85 & 70.44 & 68.15 \\
w/o PSPR
& 70.87 & 67.63 & 69.25 \\
w/o $Q_T$
& 67.59 & 71.83 & 69.71 \\
w/o $\mathcal{L}_{\mathrm{rate}}$
& 68.92 & 71.82 & \underline{70.37} \\
\midrule
\rowcolor{gray!20}
\textbf{AS-FedBridge}
& 72.21 & 71.08 & \textbf{71.64} \\
\bottomrule
\end{tabularx}
}
\caption{Additional ablations on CIFAR-100 with $\alpha=0.1$.
Bold and underline indicate the best and second-best averages.}
\label{tab:ablation}
\end{table}

AS-FedBridge achieves the highest average accuracy of 71.64\%.
Retaining only the ANN-to-SNN or SNN-to-ANN path reduces accuracy by
3.52 and 3.34 points, respectively. Globally aggregating the personalized
Bridge head causes a further 3.49-point decrease, confirming the value
of bidirectional transfer and client-specific decision boundaries.

Removing the complete Pseudo-spike Interface reduces average accuracy
by 1.81 points. Although SNN accuracy increases by 0.56 points, ANN
accuracy drops by 4.20 points, indicating that the interface mainly
improves the transfer of SNN knowledge to ANN clients. Removing PSPR,
$Q_T$, and $\mathcal{L}_{\mathrm{rate}}$ causes average decreases of
2.39, 1.93, and 1.27 points, respectively, demonstrating the
complementary roles of rate stabilization, discrete projection, and
explicit rate alignment.



\subsection{IID and Moderate Non-IID Results}
\label{app:partition_results}

Tables~\ref{tab:app_iid_results} and~\ref{tab:app_noniid_results}
compare the methods under IID and moderate non-IID partitions,
respectively. Direct parameter aggregation remains ineffective for mixed
ANN and SNN clients because their heterogeneous backbones cannot be
jointly aggregated. By contrast, methods based on transferable knowledge
provide more stable collaboration across the two client types.
AS-FedBridge maintains strong and balanced performance on both CIFAR-10
and CIFAR-100, demonstrating that the shared Bridge supports effective
ANN-SNN knowledge transfer under different data distributions. Together
with the results at $\alpha=0.1$ in the main paper, these experiments
confirm the robustness of AS-FedBridge from IID data to severe label skew.

\begin{table*}[t]
\centering
{\small
\setlength{\tabcolsep}{4.0pt}
\renewcommand{\arraystretch}{1.08}
\begin{tabular*}{\textwidth}{
@{\extracolsep{\fill}}ll*{2}{ccc}@{}
}
\toprule
\multirow{2}{*}{\textbf{Paradigm}} &
\multirow{2}{*}{\textbf{Method}} &
\multicolumn{3}{c}{\textbf{CIFAR-10}} &
\multicolumn{3}{c}{\textbf{CIFAR-100}} \\
\cmidrule(lr){3-5}\cmidrule(lr){6-8}
& & ANN & SNN & Avg. & ANN & SNN & Avg. \\
\midrule
\multirow{1}{*}{Non-FL}
& Standalone
& 73.00 & 61.90 & 67.45
& 30.90 & 12.84 & 21.87 \\
\midrule
\multirow{2}{*}{Isolated}
& Isolated FedAvg\textsuperscript{*}
& 88.26 & 88.08 & \textbf{88.17}
& 58.70 & 60.22 & \underline{59.46} \\
& Isolated FedProx\textsuperscript{*}
& 88.30 & 87.40 & \underline{87.85}
& 58.84 & 60.10 & \textbf{59.47} \\
\midrule
\multirow{3}{*}{Global-Model}
& FedAvg
& 63.40 & 9.76 & 36.58
& 28.96 & 0.80 & 14.88 \\
& FedProx
& 55.64 & 9.76 & 32.70
& 24.18 & 1.04 & 12.61 \\
& FedFree
& 76.48 & 66.54 & 71.51
& 16.24 & 30.98 & 23.61 \\
\midrule
\multirow{2}{*}{Prototype}
& FedProto
& 75.16 & 75.84 & 75.50
& 31.98 & 37.38 & 34.68 \\
& FedTGP
& 79.08 & 63.46 & 71.27
& 34.52 & 32.24 & 33.38 \\
\midrule
\multirow{3}{*}{Shared-Model}
& FML
& 79.58 & 75.16 & 77.37
& 41.50 & 37.58 & 39.54 \\
& MH-pFLID
& 74.80 & 71.32 & 73.06
& 32.38 & 32.80 & 32.59 \\
\rowcolor{gray!20}[20pt][20pt]
\cellcolor{white}
& \textbf{AS-FedBridge}
& 78.78 & 78.42 & 78.60
& 44.28 & 44.18 & 44.23 \\
\bottomrule
\end{tabular*}
}
\caption{Performance on CIFAR-10 and CIFAR-100 under IID partitioning.
The superscript \textsuperscript{*} marks architecture-isolated baselines,
which aggregate ANN and SNN clients separately. Since IID clients follow
the same data distribution, these baselines approximate standard
distributed training within each architecture. Bold and underline
indicate the best and second-best average results.}
\label{tab:app_iid_results}
\end{table*}

\begin{table*}[t]
\centering
{\small
\setlength{\tabcolsep}{4.0pt}
\renewcommand{\arraystretch}{1.08}
\begin{tabular*}{\textwidth}{
@{\extracolsep{\fill}}ll*{2}{ccc}@{}
}
\toprule
\multirow{2}{*}{\textbf{Paradigm}} &
\multirow{2}{*}{\textbf{Method}} &
\multicolumn{3}{c}{\textbf{CIFAR-10}} &
\multicolumn{3}{c}{\textbf{CIFAR-100}} \\
\cmidrule(lr){3-5}\cmidrule(lr){6-8}
& & ANN & SNN & Avg. & ANN & SNN & Avg. \\
\midrule

\multirow{1}{*}{Non-FL}
& Standalone
& 79.71 & 75.69 & 77.70
& 52.36 & 52.89 & 52.62 \\
\midrule

\multirow{2}{*}{Isolated}
& Isolated FedAvg
& 82.86 & 89.59 & \underline{86.23}
& 60.09  & 53.75 & \underline{56.92} \\
& Isolated FedProx
& 83.48 & 85.23 & 84.36
& 60.85 & 51.02 & 55.94 \\
\midrule

\multirow{3}{*}{Global-Model}
& FedAvg
& 62.69 & 19.20 & 40.95
& 25.87 & 1.64 & 13.76 \\
& FedProx
& 59.27 & 13.49 & 36.38
& 20.22 & 0.45 & 10.34 \\
& FedFree
& 70.52 & 75.38 & 72.95
& 44.31 & 28.84 & 36.57 \\
\midrule

\multirow{2}{*}{Prototype}
& FedProto
& 80.21 & 85.18 & 82.70
& 50.33 & 54.19 & 52.26 \\
& FedTGP
& 81.06 & 76.87 & 78.97
& 52.57 & 44.11 & 48.34 \\
\midrule

\multirow{3}{*}{Shared-Model}
& FML
& 82.86 & 86.75 & 84.81
& 55.33 & 53.49 & 54.41 \\
& MH-pFLID
& 80.73 & 78.45 & 79.59
& 56.88 & 50.84 & 53.86 \\
\rowcolor{gray!20}[20pt][20pt]
\cellcolor{white}
& \textbf{AS-FedBridge}
& 87.94 & 85.76 & \textbf{86.85}
& 59.20 & 56.54 & \textbf{57.87} \\
\bottomrule
\end{tabular*}
}
\caption{Performance on CIFAR-10 and CIFAR-100 under Dirichlet non-IID
partitioning with $\alpha=0.5$. Bold and underline indicate the best and
second-best results.}
\label{tab:app_noniid_results}
\end{table*}


\subsection{Maximum Mean Discrepancy Analysis}
\label{app:mmd_results}

We use squared Maximum Mean Discrepancy
($\operatorname{MMD}^{2}$) to evaluate the feature-distribution gap
between ANN and SNN clients. For each method trained under the IID
setting on CIFAR-10 and CIFAR-100, the five ANN clients and five SNN
clients process the same fixed probe set of $M=1{,}000$ test images from
the corresponding dataset. We extract the final ANN feature $h_a(x)$
before the classifier and temporally average the corresponding SNN
feature:
\begin{equation}
\bar h_s(x)
=
\frac{1}{T}
\sum_{t=1}^{T}
h_s^t(x).
\label{eq:app_mmd_snn_feature}
\end{equation}
Because this analysis uses ResNet-18 and Spiking ResNet-18, the extracted
features have the same dimension. The complete probe distributions are
compared without class-wise grouping, learned projection, or post-hoc
normalization.

We use the RBF kernel
$k(u,v)=\exp[-\|u-v\|_2^2/(2\sigma_k^2)]$ and approximate its mean
embedding with $D=1024$ random Fourier features:
\begin{equation}
\begin{gathered}
z(u)
=
\sqrt{\frac{2}{D}}
\cos\!\left(W^\top u+b\right),\\
\widehat{\mu}_{a}
=
\frac{1}{M}
\sum_{q=1}^{M}
z\!\left(h_a(x_q)\right),\\
\widehat{\mu}_{s}
=
\frac{1}{M}
\sum_{q=1}^{M}
z\!\left(\bar h_s(x_q)\right),\\
\widehat{\operatorname{MMD}}_{a,s}^{2}
=
\left\|
\widehat{\mu}_{a}
-
\widehat{\mu}_{s}
\right\|_2^2.
\end{gathered}
\label{eq:app_mmd_rff}
\end{equation}
The random seed is fixed to $42$. For each dataset, the kernel bandwidth
is selected once using the median heuristic on a deterministic pooled
subsample of at most $512$ probe features and is then fixed across all
methods and client pairs.

Let
$\mathcal Q=
\mathcal C_{\mathrm{ann}}\times\mathcal C_{\mathrm{snn}}$
denote the set of all $25$ ANN and SNN client pairs. The method-level
distribution discrepancy is
\begin{equation}
\overline{\operatorname{MMD}^{2}}
=
\frac{1}{|\mathcal Q|}
\sum_{(a,s)\in\mathcal Q}
\widehat{\operatorname{MMD}}_{a,s}^{2}.
\label{eq:app_pairwise_mmd}
\end{equation}
For the pairwise matrices, the same estimator is applied to every pair
among the ten clients, including the ANN, SNN, and cross-type blocks.

\begin{figure*}[t]
\centering
\includegraphics[width=0.95\textwidth]{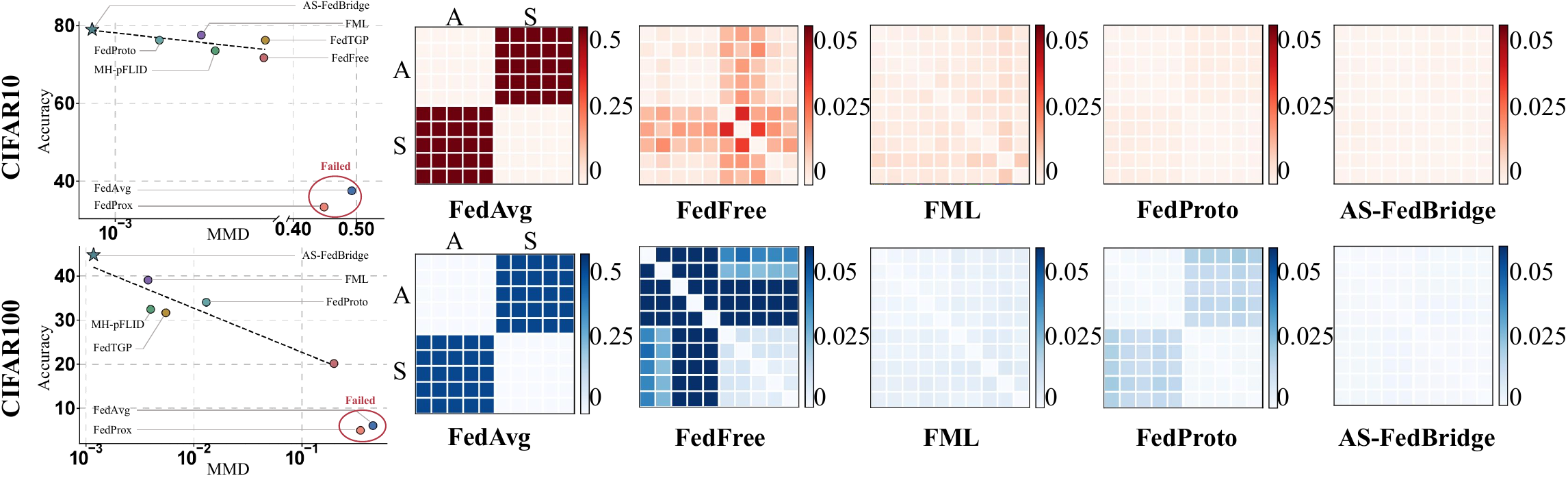}
\caption{Feature-distribution alignment measured by Maximum Mean
Discrepancy (MMD) and collaborative accuracy on IID CIFAR-10 and
CIFAR-100. AS-FedBridge substantially reduces the distribution gap
between continuous and spiking clients while achieving the highest
accuracy, showing that the shared Bridge learns a compatible
representation space for mixed collaboration.}
\label{fig:app_mmd}
\end{figure*}

Figure~\ref{fig:app_mmd} shows a consistent negative association between
the ANN and SNN distribution discrepancy and collaborative accuracy on
both datasets. Direct FedAvg and FedProx produce large cross-type
$\operatorname{MMD}^{2}$ values and severe performance degradation. In
contrast, AS-FedBridge achieves the smallest distribution gap and the
highest accuracy. The pairwise matrices further show that AS-FedBridge
substantially weakens the block separation between ANN and SNN clients.
These results indicate that the shared Bridge makes continuous and
spiking representations statistically more compatible without
sacrificing their discriminative ability.
$\operatorname{MMD}^{2}$ is used only as a post-training diagnostic and
is not included in the training objective.


\subsection{Sensitivity to Bridge Width}
\label{app:bridge_sensitivity}

We investigate the influence of Bridge capacity by uniformly scaling the
width of its shared body to $0.5\times$, $0.75\times$, and $1.0\times$.
All variants use the same private ANN and SNN backbones, data partition,
and training configuration. This comparison isolates the effect of Bridge
size while reflecting its trade-off with communication and local
computation costs.

\begin{table}[t]
\centering
{\small
\setlength{\tabcolsep}{4.5pt}
\renewcommand{\arraystretch}{1.10}
\begin{tabularx}{\columnwidth}{
@{}>{\raggedright\arraybackslash}Xccc@{}
}
\toprule
\textbf{Bridge Width} & \textbf{ANN} & \textbf{SNN} &
\textbf{Avg.} \\
\midrule
$0.5\times$  & 71.38 & 67.41 & 69.40 \\
$0.75\times$ & 72.07 & 68.26 & \underline{70.17} \\
\rowcolor{gray!20}
$1.0\times$ (default)
& 72.21 & 71.08 & \textbf{71.64} \\
\bottomrule
\end{tabularx}
}
\caption{Sensitivity of AS-FedBridge to Bridge width on CIFAR-100 with
Dirichlet non-IID partitioning ($\alpha=0.1$).}
\label{tab:app_bridge_sensitivity}
\end{table}

Table~\ref{tab:app_bridge_sensitivity} shows that increasing the Bridge
width consistently improves performance. Reducing the width from
$1.0\times$ to $0.75\times$ and $0.5\times$ decreases the average
accuracy from 71.64\% to 70.17\% and 69.40\%, respectively. ANN accuracy
changes moderately from 72.21\% to 72.07\% and 71.38\%, whereas SNN
accuracy is more sensitive, decreasing from 71.08\% to 68.26\% and
67.41\%. We therefore use the $1.0\times$ Bridge by default, while the
narrower variants provide compact alternatives for resource-constrained
clients.

\subsection{Comparison with Multi-Bit Quantized Networks}
\label{app:qnn_results}

We extend the binary-network comparison in the main paper to 2-, 4-,
and 8-bit Quantized Neural Networks (QNNs). The homogeneous setting
trains ten QNN clients using FedAvg, while the mixed setting contains
five full-precision ANN clients and five QNN clients connected through
the same Bridge. We also report homogeneous BNN, SNN, and ANN controls,
together with mixed configurations pairing ANN clients with SNN or ANN
clients. All experiments use CIFAR-100 with Dirichlet non-IID
partitioning at $\alpha=0.1$ and share the same client data, backbone
family, communication rounds, and number of local epochs.

Following the binary-network setting, the eligible convolutional weights
and activations are quantized, while the input and output layers remain
in full precision. These full-precision layers account for 3.85M FLOPs.
Since one dense multiply-accumulate operation corresponds to two FLOPs,
the remaining quantized portion contains
\[
N_{\mathrm{MAC}}^{\mathrm{low}}
=
\frac{1111.15-3.85}{2}
=
553.65\text{M}
\]
low-bit multiply-accumulate operations. In the mixed setting, low-bit
clients constitute half of the federation, yielding approximately
276.82M low-bit operations per client on average. This operation count
is unchanged across 2-, 4-, and 8-bit QNNs because their topology is
identical. Their BitOP costs differ according to
$N_{\mathrm{MAC}}^{\mathrm{low}}b_wb_a$, reflecting the weight and
activation precisions. The mixed-setting FLOPs additionally include
the full-precision computation of the ANN clients and the Bridge.

\begin{table}[!t]
\centering
\setlength{\abovecaptionskip}{3pt}
\setlength{\belowcaptionskip}{0pt}
{\small
\setlength{\tabcolsep}{1.2pt}
\renewcommand{\arraystretch}{0.96}
\begin{tabularx}{\columnwidth}{
@{}>{\raggedright\arraybackslash}Xlcccc@{}
}
\toprule
& & &
\multicolumn{3}{c}{\textbf{Computation (M)}} \\
\cmidrule(lr){4-6}
\textbf{Setting} &
\textbf{Method} &
\textbf{Acc.} $\uparrow$ &
\textbf{FLOPs} $\downarrow$ &
\textbf{BitOPs} $\downarrow$ &
\textbf{SOPs} $\downarrow$ \\
\midrule

\multicolumn{6}{@{}l}{\textit{Homogeneous Collaboration}} \\

BNN (1-bit)
& FedAvg
& 41.25 \textsuperscript{*}
& 3.85$^{\dagger}$
& 553.65
& 0 \\

QNN (2-bit)
& FedAvg
& 53.35 \textsuperscript{*}
& 3.85$^{\dagger}$
& 2214.60
& 0 \\

QNN (4-bit)
& FedAvg
& 61.52 \textsuperscript{*}
& 3.85$^{\dagger}$
& 8858.40
& 0 \\

QNN (8-bit)
& FedAvg
& 62.31 \textsuperscript{*}
& 3.85$^{\dagger}$
& 35433.60
& 0 \\

SNN ($T=2$)
& FedAvg
& 59.39
& 0
& 0
& 81.69 \\

SNN ($T=4$)
& FedAvg
& 61.06
& 0
& 0
& 167.17 \\

SNN ($T=8$)
& FedAvg
& 61.81
& 0
& 0
& 338.29 \\

ANN
& FedAvg
& 63.98 \textsuperscript{*}
& 1111.15
& 0
& 0 \\

\midrule
\multicolumn{6}{@{}l}{\textit{Mixed Collaboration}} \\

ANN and BNN (1-bit)
& Bridge
& 60.84 \textsuperscript{*}
& 655.59$^{\ddagger}$
& 276.82
& 0 \\

ANN and QNN (2-bit)
& Bridge
& 65.04 \textsuperscript{*}
& 655.59$^{\ddagger}$
& 1107.30
& 0 \\

ANN and QNN (4-bit)
& Bridge
& 67.09 \textsuperscript{*}
& 655.59$^{\ddagger}$
& 4429.20
& 0 \\

ANN and QNN (8-bit)
& Bridge
& 70.48 \textsuperscript{*}
& 655.59$^{\ddagger}$
& 17716.80
& 0 \\

\rowcolor{gray!20}
\textbf{ANN and SNN ($T=2$)}
& \textbf{AS-FedBridge}
& 67.01
& 653.67
& 0
& 40.85 \\

\rowcolor{gray!20}
\textbf{ANN and SNN ($T=4$)}
& \textbf{AS-FedBridge}
& 71.64
& 653.67
& 0
& 83.58 \\

\rowcolor{gray!20}
\textbf{ANN and SNN ($T=8$)}
& \textbf{AS-FedBridge}
& 72.49
& 653.67
& 0
& 169.15 \\

ANN and ANN
& Bridge
& 72.67 \textsuperscript{*}
& 1209.24
& 0
& 0 \\

\bottomrule
\end{tabularx}
}
\caption{Performance and computation costs on CIFAR-100 with
$\alpha=0.1$. \textsuperscript{*} marks models still require
floating-point computation. $^\dagger$ and $^\ddagger$ indicate
553.65M and 276.82M additional low-bit MACs, respectively. The 1-bit
case can be implemented using XNOR-popcount operations. BitOPs are
computed as $N_{\mathrm{MAC}}^{\mathrm{low}}b_wb_a$, where $b_w$ and
$b_a$ denote the weight and activation bit widths.}
\label{tab:app_qnn}
\end{table}

Table~\ref{tab:app_qnn} separates the effects of arithmetic precision
and temporal spike processing. In the homogeneous setting, QNN accuracy
increases from 53.35\% at 2 bits to 62.31\% at 8 bits, while SNN accuracy
increases from 59.39\% at $T=2$ to 61.81\% at $T=8$. Increasing QNN
precision in the mixed setting improves accuracy from 65.04\% to
70.48\%.

More importantly, AS-FedBridge consistently performs better under
architecture heterogeneity. For the corresponding 2, 4, and 8
configurations, it outperforms the mixed ANN and QNN baselines by
1.97, 4.55, and 2.01 percentage points, respectively. AS-FedBridge
with $T=4$ already exceeds the strongest 8-bit QNN control by 1.16
points. Increasing the time steps to $T=8$ further raises accuracy to
72.49\%, only 0.18 points below the ANN and ANN control, while requiring
653.67M FP32 FLOPs compared with 1209.24M. The additional temporal cost
increases from 40.85M SOPs at $T=2$ to 169.15M at $T=8$. We report
FP32 FLOPs, BitOPs, and SOPs separately because their relative hardware
costs depend on the target platform.

\section{Experimental Details}
\label{app:protocol}

This section provides the implementation details used throughout our
experiments. We first describe the common settings shared by all
experiments, including the federated configuration, data partitions, data
processing, and optimization. We then explain how existing federated
learning methods are adapted to the mixed ANN and SNN setting and how ANN
and SNN representations are compared. Finally, we present the specific
settings for the motivation, mechanism, main-result, necessity, ablation,
communication, and computation experiments. We provide full parameters in Table~\ref{tab:app_default_parameters}.

\subsection{Common Experimental Settings}
\label{app:common_protocol}

\paragraph{Federated configuration.}
We consider a federation of $N$ clients, divided into ANN clients
$\mathcal{C}_{\rm ann}$ and SNN clients $\mathcal{C}_{\rm snn}$.
Each client $i$ holds a private training set $\mathcal{D}_i$, a disjoint
local test set $\mathcal{D}_i^{\rm te}$, and a private backbone
$f_i(\cdot;\theta_i)$. AS-FedBridge maintains a shared Bridge body
$b(\cdot;\omega)$ and a personalized classifier head
$g_i(\cdot;\psi_i)$ for each client. Only the shared body $\omega$ is
uploaded and aggregated, while the local datasets, private backbones, and
personalized heads remain on the clients.

Unless otherwise specified, all experiments contain ten fully
participating clients, comprising five ResNet-18 ANN clients and five
Spiking-ResNet18 SNN clients. SNN clients use $T=4$ time steps by
default, with their states reset between independent samples. For static
images, the same input is presented at each time step, and SNN predictions
are obtained by averaging the temporal outputs. We use $\bar{Z}_i$ and
$\bar{F}_i$ to denote the temporally averaged logits and final features,
respectively.

\paragraph{Data partitions.}
For IID experiments, independent shuffled permutations of the training and
test sets are divided into ten nearly equal shards. For a Dirichlet partition,
the client proportions are sampled independently for each class $c$,
\begin{equation}
  \boldsymbol p_c\sim\operatorname{Dirichlet}(\alpha\mathbf 1).
  \label{eq:app_dirichlet}
\end{equation}
The training allocation is redrawn until every client has more than ten
examples. Test examples are then assigned class by class to reproduce the
training class mixture of each client. Thus, the local test distribution is
matched to its training shard instead of being generated by an unrelated
second Dirichlet draw. We use $\alpha=0.1$ for severe label skew and
$\alpha=0.5$ for the moderate non-IID supplement. 

\paragraph{Data processing.}
For CIFAR-10 and CIFAR-100, training images are randomly cropped to
$32\times32$ with four-pixel padding and horizontally flipped. Images are
normalized using mean $(0.4914,0.4822,0.4465)$ and standard deviation
$(0.2470,0.2435,0.2616)$. Evaluation applies only normalization. For
Tiny-ImageNet, training uses random resized crops to $224\times224$ and
horizontal flipping. Evaluation images are resized to 256 and center-cropped
to $224\times224$, followed by standard ImageNet normalization.

For CIFAR10-DVS, each event sample is divided into $T=10$ frames by event
count and resized to $48\times48$. The same spatial augmentation is applied
across all time steps and polarities to preserve temporal consistency. SNN
clients directly process the resulting frames, while ANN clients fold the
temporal and polarity dimensions into the input channels. Both client types
therefore use the same event representation and data partition.

\paragraph{Optimization.}
All experiments run for 100 communication rounds. In each AS-FedBridge
round, the private backbone is updated for five local epochs during
extraction, followed by one local epoch of Bridge injection. The training and evaluation batch sizes are both
256, and the random seed is fixed to 42. We use SGD with momentum 0.9 and
cosine learning-rate decay for all local models. ANN backbones use an initial
learning rate of 0.05 and weight decay of $10^{-4}$, while SNN backbones use
an initial learning rate of 0.1 and weight decay of $5\times10^{-4}$. The
Bridge is optimized with an initial learning rate of 0.004. Biases and
normalization parameters are excluded from weight decay.

All clients participate in each communication round. After local training,
the server aggregates the shared Bridge body according to the size of each
client's training set:
\begin{equation}
  \omega^{r+1}
  =
  \sum_{i\in\mathcal S_r}
  \frac{n_i}
  {\sum_{j\in\mathcal S_r}n_j}
  \omega_i^{r+1},
  \label{eq:app_aggregation}
\end{equation}
where $\mathcal S_r$ denotes the participating clients in round $r$, and
$n_i$ is the number of training samples held by client $i$. Private ANN and
SNN backbones are evaluated locally and are never included in server
aggregation.

\begin{table}[t]
\centering
{\small
\setlength{\tabcolsep}{3.5pt}
\renewcommand{\arraystretch}{1.08}
\begin{tabularx}{\columnwidth}{
@{}>{\raggedright\arraybackslash}X
>{\raggedright\arraybackslash}X@{}
}
\toprule
\textbf{Parameter} & \textbf{Default Setting} \\
\midrule

\multicolumn{2}{@{}l}{\textit{Federated Training}} \\

Client configuration
& $N=10$ with 5 ANN and 5 SNN clients \\

Data partition
& Dirichlet non-IID, $\alpha=0.1$ \\

Participation and rounds
& Full participation, $R=100$ \\

Local epochs
& $E_{\mathrm{ext}}=5$, $E_{\mathrm{inj}}=1$ \\

Batch size
& 256 for training and evaluation \\

SNN time steps
& $T=4$; $T=10$ for CIFAR10-DVS \\

Optimizer and schedule
& SGD, momentum $0.9$, cosine decay \\

Initial learning rate
& ANN: $0.05$; SNN: $0.1$; Bridge: $0.004$ \\

Weight decay
& ANN: $10^{-4}$; SNN: $5\times10^{-4}$ \\

Random seed
& 42 \\

\midrule
\multicolumn{2}{@{}l}{\textit{AS-FedBridge Parameters}} \\

Distillation
& $\tau=2.0$,
  $\alpha_{kd}=0.10\rightarrow0.025$ (ANN);
  $0.07\rightarrow0.025$ (SNN) \\

Bridge injection
& $\alpha_{teach}=0.16\rightarrow0.05$,
  $\alpha_{ce}=1.10\rightarrow0.55$ \\

Rate alignment
& $\alpha_{rate}=0.005$,
  $\alpha_{pspr}=0.10$ \\

Proximal regularization
& $\alpha_{prox}=10^{-4}$ \\

SNN extraction
& $\lambda_{nld}=0.08$,
  $\lambda_{tet}=10^{-4}$ \\

PSPR mean constraint
& $V_{\mathrm{th}}=0.5$,
  $k_d=1.0$ \\

PSPR variance constraint
& $\sigma_{\min}=0.05$,
  $\lambda_{\mathrm{var}}=1.0$ \\

Bridge configuration
& $1.0\times$ width; 3.93 MB in FP32 \\

\bottomrule
\end{tabularx}
}
\caption{Default training and method-specific parameters of
AS-FedBridge. $E_{\mathrm{ext}}$ and $E_{\mathrm{inj}}$ denote the
numbers of local extraction and injection epochs.}
\label{tab:app_default_parameters}
\end{table}

\subsection{How Existing FL Methods Are Adapted}
\label{app:baseline_adaptation}

\begin{table*}[t]
\centering
{\small
\setlength{\tabcolsep}{2.2pt}
\renewcommand{\arraystretch}{1.10}
\begin{tabularx}{\textwidth}{
@{}ll
>{\raggedright\arraybackslash}X
>{\raggedright\arraybackslash}X
>{\raggedright\arraybackslash}X@{}
}
\toprule
\textbf{Method Type} &
\textbf{Method} &
\textbf{Communicated Object} &
\textbf{Adaptation for ANN and SNN Clients} &
\textbf{SNN Processing} \\
\midrule

\multirow{2}{*}{Separate}
& Isolated FedAvg
& Type-specific ANN and SNN backbones
& Maintains separate aggregates for ANN and SNN clients without transferring knowledge between them.
& Uses temporally averaged logits for prediction. \\

& Isolated FedProx
& Type-specific ANN and SNN backbones
& Applies the original proximal objective independently within each client type.
& Applies the proximal term only to persistent trainable parameters. \\

\midrule
\multirow{2}{*}{Direct}
& FedAvg
& Full compatible backbone
& Directly aggregates corresponding trainable tensors when the ANN and SNN backbones have compatible shapes.
& Averages temporal logits for supervision and prediction. \\

& FedProx
& Full compatible backbone
& Uses the same parameter mapping as FedAvg and adds the original proximal objective.
& Excludes spikes, membrane states, and other temporal states from aggregation. \\

\midrule
Layer selection
& FedFree
& Selected critical layers
& Retains the original critical-layer selection, layer-wise transfer, and response alignment.
& Uses temporally averaged responses when static supervision is required. \\

\midrule
\multirow{2}{*}{Prototype}
& FedProto
& Class-wise prototypes
& Constructs ANN and SNN prototypes in the same feature dimension and applies the original prototype aggregation.
& Uses the temporally averaged final feature $\bar h_i$. \\

& FedTGP
& Trainable global prototypes
& Uses the same client-side prototype construction as FedProto and retains its original server optimization.
& Uses $\bar h_i$ as the SNN prototype representation. \\

\midrule
\multirow{3}{*}{Global model}
& FML
& Shared distillation model
& Retains mutual learning and aggregates a size-matched lightweight model.
& Uses temporally averaged logits in the mutual-learning objective. \\

& MH-pFLID
& Lightweight messenger
& Retains its original injection, distillation, and personalization rules with the same model budget.
& Uses temporally averaged logits and features when required. \\

\rowcolor{gray!20}
& \textbf{AS-FedBridge}
& \textbf{Shared Bridge body}
& \textbf{Uses bidirectional distillation while keeping each backbone and classifier head private.}
& \textbf{Uses temporal logits and bottleneck firing rates during SNN training.} \\

\bottomrule
\end{tabularx}
}
\caption{Adaptation of existing FL methods to the mixed ANN and SNN setting.
FML, MH-pFLID, and AS-FedBridge use the same communicated model budget.}
\label{tab:app_baseline_adaptation}
\end{table*}

Table~\ref{tab:app_baseline_adaptation} summarizes how each baseline is
implemented for ANN and SNN clients. All methods use the same client
identities, data partitions, private backbones, backbone initializations,
communication rounds, batch size, and evaluation protocol. Each private
backbone receives the same five-epoch local training budget. Method-specific
objectives and server updates follow their original formulations.

For methods that communicate a lightweight model, we use the same architecture
and communicated parameter budget for the FML distillation model, the
MH-pFLID messenger, and the AS-FedBridge body. Each communicated component
occupies approximately 4\,MB in FP32 and is uploaded and downloaded once per
communication round. Their initialization protocol and optimization budget are
also matched. This setting controls both communication capacity and model size,
so the comparison reflects differences in the learning objectives and update
mechanisms. The Pseudo-Spike Interface, rate alignment, Pseudo-Spike
Polarization Regularization, and bidirectional Bridge training are used only
by AS-FedBridge and are not added to any baseline.

\paragraph{Isolated FedAvg and FedProx.}
These variants maintain independent global backbones for ANN and SNN clients.
Aggregation is performed only among clients of the same type, so they measure
the performance obtained without knowledge transfer between ANN and SNN
clients. FedProx additionally applies its original proximal term within each
group.

\paragraph{FedAvg and FedProx.}
For the ResNet-18 and Spiking-ResNet18 setting, the two backbones contain
compatible trainable convolutional and classifier tensors. FedAvg directly
aggregates the corresponding tensors across all clients after local supervised
training. Spikes, membrane potentials, and other transient neuronal states
remain local. FedProx uses the same parameter mapping and adds its original
proximal penalty. These adaptations provide a direct evaluation of shared
parameter aggregation between ANN and SNN clients.

When backbone widths or architectures differ, direct aggregation is reported
only if the baseline defines a valid parameter mapping. Incompatible tensors
are not padded, truncated, or transformed by an additional network.

\paragraph{FedFree.}
FedFree retains its original critical-layer selection and response-based
transfer. Only persistent trainable convolutional and linear parameters are
eligible for communication. SNN states are excluded. Its layer selection,
transfer order, and server update follow the original method.

\paragraph{FedProto and FedTGP.}
Both methods construct class prototypes from the final feature before the
classifier. For an SNN client, the feature is averaged over time as
\begin{equation}
  \bar h_i
  =
  \frac{1}{T}\sum_{t=1}^{T}h_i^t.
  \label{eq:app_snn_feature_mean}
\end{equation}
FedProto and FedTGP use the same prototype dimension and local projection rule
when their native feature dimensions differ. A class prototype is computed
only from locally observed examples, while absent classes are excluded from
aggregation. FedProto averages the resulting client prototypes, whereas
FedTGP retains its trainable global prototype generator and original server
objective.

\paragraph{FML and MH-pFLID.}
FML jointly optimizes each private backbone and a lightweight distillation
model through mutual learning. MH-pFLID retains its messenger-based injection,
distillation, and personalization procedures. Their communicated models use
the same architecture and parameter count as the communicated Bridge body in
AS-FedBridge. Only the output heads are adjusted to the number of dataset
classes. When a static SNN representation is required, both methods use the
temporal averages
\begin{equation}
  \bar Z_i
  =
  \frac{1}{T}\sum_{t=1}^{T}Z_i^t,
  \qquad
  \bar h_i
  =
  \frac{1}{T}\sum_{t=1}^{T}h_i^t.
\end{equation}
The private ANN or SNN backbone remains the test-time predictor for every
method.

\paragraph{Standalone training.}
Standalone clients use the same private backbones, data partitions,
initializations, optimizers, and local training budget but do not communicate.
This setting serves as a no-collaboration reference rather than a centralized
upper bound.

\subsection{Bridge and SNN Feature Alignment}
\label{app:one_to_one_alignment}

AS-FedBridge does not require layer-wise correspondence between different
backbones. Instead, rate alignment is performed only at a designated Bridge
bottleneck $\ell^\star$. This design avoids matching intermediate layers whose
depth, spatial resolution, and channel dimension may differ across ResNet,
VGG, and other architectures.

For SNN client $i$, let $H_i^t(x)$ denote the final spiking feature at time
step $t$. Global average pooling first removes the spatial dimension. When
the resulting channel dimension differs from that of the Bridge bottleneck,
a private spiking projector $p_i$ maps it to a fixed dimension $d$:
\begin{equation}
  S_i^t(x)
  =
  p_i\!\left(\operatorname{GAP}\!\left(H_i^t(x)\right)\right)
  \in\{0,1\}^{d}.
  \label{eq:app_snn_projection}
\end{equation}
The projector is replaced by an identity mapping when the dimensions already
match. It is trained locally with the SNN backbone during extraction, remains
private, and is frozen during Bridge injection. The corresponding firing-rate
representation is
\begin{equation}
  r_i(x)
  =
  \frac{1}{T}\sum_{t=1}^{T}S_i^t(x).
  \label{eq:app_snn_rate}
\end{equation}

For the same input $x$, the Bridge produces a pseudo-spike representation
$\widetilde r_B^{\ell^\star}(x)\in
\{0,1/T,\ldots,1\}^{d}$ at bottleneck $\ell^\star$. The Bridge and SNN
representations therefore have the same sample order, dimension, and
rate support. For a batch containing $B$ samples, rate alignment is defined as
\begin{equation}
  \mathcal L_{\mathrm{rate}}^{i}
  =
  \frac{1}{Bd}
  \sum_{b=1}^{B}\sum_{q=1}^{d}
  \left(
  \widetilde r_{B,bq}^{\ell^\star}
  -
  \sg\!\left[r_{i,bq}\right]
  \right)^2.
  \label{eq:app_coordinate_alignment}
\end{equation}
The stop-gradient operator keeps the SNN representation fixed during
injection, so this objective updates only the local Bridge. Repeated
sample-wise matching over local batches encourages the empirical Bridge rate
distribution to follow the firing-rate distribution observed by each SNN
client. The finite-rate projection defines their common support, while
Pseudo-Spike Polarization Regularization prevents the Bridge representation
from collapsing to a narrow range.

ANN and SNN backbones are therefore not directly matched at arbitrary
intermediate layers. ANN clients transfer knowledge to the continuous Bridge
through logit distillation, while SNN clients additionally calibrate the
Bridge bottleneck through rate alignment. Server aggregation combines these
updates in the shared Bridge body, providing a common exchange representation
without requiring identical backbone architectures or native feature
dimensions.

\subsection{Motivation and Mechanism Analysis}
\label{app:motivation_protocol}

\paragraph{Direct FedAvg motivation.}
Although ResNet-18 and Spiking ResNet-18 use different forward dynamics,
their persistent trainable tensors have matching shapes. We therefore
apply FedAvg directly to examine whether parameter compatibility alone
enables ANN and SNN collaboration. The experiment is trained from random
initialization under a non-IID partition of CIFAR-10 and CIFAR-100, with five
ANN clients and five SNN clients. All clients participate in every round
and perform five local epochs before data-size-weighted aggregation.
SNN membrane potentials, accumulated spikes, and other temporal states
remain local. No Bridge, distillation loss, or alignment component is
used. The ANN-only and SNN-only references follow the same partition,
initialization, optimization, and training budgets.

After each local training stage and before aggregation, all client models
process the same ordered probe set of $M=1{,}000$ test images. We extract
the final ANN feature and logits directly, while the SNN outputs are
averaged over time:
\begin{equation}
\begin{aligned}
\bar h_i(x)
&=
\frac{1}{T}\sum_{t=1}^{T}h_i^t(x),\\
\bar Z_i(x)
&=
\frac{1}{T}\sum_{t=1}^{T}Z_i^t(x).
\end{aligned}
\label{eq:app_diagnostic_snn_average}
\end{equation}
The probe set is used only for diagnosis and does not affect training or
aggregation. Pairwise CKA, $\operatorname{MMD}^{2}$, and gradient cosine
similarity are averaged over the $10$ ANN pairs, $25$ ANN and SNN pairs,
and $10$ SNN pairs. The resulting low cross-type CKA, large
$\operatorname{MMD}^{2}$, and low gradient similarity explain why direct
FedAvg causes the performance collapse reported in the motivation figure.

The round-wise $\operatorname{MMD}^{2}$ uses the estimator defined in
Section~\ref{app:mmd_results}. It differs from the additional experiment
in that section, which compares the final checkpoints of all methods and
reports only the $25$ ANN and SNN pairs.

\paragraph{CKA and gradient consistency.}
\label{app:mechanism_protocol}

For the mechanism comparison, each method is first trained to completion
under the same IID setting. Its final client models are then frozen and
evaluated on the common probe set. For centered feature matrices
$X\in\mathbb R^{M\times d_x}$ and
$Y\in\mathbb R^{M\times d_y}$, linear CKA is
\begin{equation}
\operatorname{CKA}(X,Y)
=
\frac{\|X^\top Y\|_F^2}
{\|X^\top X\|_F\,
 \|Y^\top Y\|_F}.
\label{eq:app_cka}
\end{equation}
Centering is performed over the probe examples. No learned projection or
post-hoc transformation is introduced. We compute the complete pairwise
CKA matrix within each method and report the ANN, SNN, and cross-type
blocks shown in the main paper.

Gradient consistency is measured on the same probe set using only the
supervised classification loss. Let $\theta_i^{\mathrm{cmp}}$ denote the
persistent convolutional and classifier parameters shared by the
ResNet-18 and Spiking ResNet-18 implementations. We compute
\begin{equation}
\begin{aligned}
\ell_i
&=
\frac{1}{M}
\sum_{m=1}^{M}
\mathcal L_{\mathrm{CE}}
\left(\widehat Z_i(x_m),y_m\right),\\
g_i
&=
\operatorname{vec}\left(
\nabla_{\theta_i^{\mathrm{cmp}}}\ell_i
\right),
\end{aligned}
\label{eq:app_probe_gradient}
\end{equation}
where $\widehat Z_i=Z_i$ for an ANN client and
$\widehat Z_i=\bar Z_i$ for an SNN client. Pairwise gradient similarity is
\begin{equation}
\operatorname{Cos}(i,j)
=
\frac{\langle g_i,g_j\rangle}
{\|g_i\|_2\|g_j\|_2+\epsilon}.
\label{eq:app_gradient_cosine}
\end{equation}
Method-specific training losses are excluded from this diagnostic, and
the computed gradients are not applied to the models. Cosines are
compared only between clients within the same method. The near-zero or
negative ANN and SNN similarities produced by several baselines indicate
conflicting optimization directions, whereas AS-FedBridge yields broadly
positive cross-type similarities.

\paragraph{Pseudo-spike and SNN distributions.}

The distribution analysis in the main paper is performed after training
AS-FedBridge. We evaluate the local Bridge and frozen SNN of non-IID
client 9 on the same local test shard of $640$ examples with $T=4$.
At each of the four residual stages and the output interface, we collect
the quantized Bridge pseudo-spike activation
$\widetilde r_B^\ell$ and the temporally averaged SNN firing rate
\begin{equation}
\bar F_i^\ell
=
\frac{1}{T}
\sum_{t=1}^{T}F_i^{t,\ell}.
\label{eq:app_stage_rate}
\end{equation}
For either
$R^\ell=\widetilde r_B^\ell$ or
$R^\ell=\bar F_i^\ell$, all examples and feature coordinates are
flattened into $N_\ell$ values. The independently normalized histogram is
\begin{equation}
\widehat p_{R,\ell}(k)
=
\frac{1}{N_\ell}
\sum_{u=1}^{N_\ell}
\mathbf 1
\left[
R_u^\ell=\frac{k}{T}
\right],
\label{eq:app_rate_histogram}
\end{equation}
where $k=0,\ldots,T$. Both distributions therefore share the rate support
$\mathcal R_4=\{0,0.25,0.5,0.75,1\}$.

This comparison is restricted to the homogeneous setting, where the four
residual stages have corresponding semantic positions. The stage-wise
histograms are post-training visualizations rather than additional
alignment losses. During training, rate matching is applied only at the
designated Bridge bottleneck $\ell^\star$. The changing probability mass
across stages shows that the Bridge captures layer-dependent firing
patterns instead of collapsing to a fixed discrete distribution.

\subsection{Main-Result Experiment}
\label{app:main_results_protocol}

\paragraph{Common settings.}
All main-result experiments use ten clients with full participation and
Dirichlet label skew at $\alpha=0.1$. For each class $c$, its allocation
across clients is sampled as
\begin{equation}
\boldsymbol{\pi}_c
\sim
\operatorname{Dirichlet}
\left(0.1\,\mathbf 1_{10}\right),
\qquad c=1,\ldots,C.
\label{eq:app_main_dirichlet}
\end{equation}
For each dataset, the partition and client order are generated once and
reused by every method. All methods use the same private-backbone initializations, communication
rounds, batch size, evaluation frequency, and five-epoch
private-backbone update budget per round. AS-FedBridge additionally
performs one Bridge-injection epoch, whose cost is included in the
computation analysis. Baselines follow the
adaptations in Section~\ref{app:baseline_adaptation}. The private ANN or
SNN backbone, rather than an auxiliary communication model, is always
used for evaluation.

\paragraph{Homogeneous backbone setting.}
The basic mixed setting contains five full-width ResNet-18 ANN clients
and five full-width Spiking ResNet-18 clients. It isolates the
representation difference between continuous and spiking models while
keeping the architecture and model capacity fixed within each client
type. This setting is evaluated on CIFAR-10, CIFAR-100, Tiny-ImageNet,
and CIFAR10-DVS. Centralized and standalone training provide non-FL
references, while isolated FedAvg and FedProx aggregate the ANN and SNN
groups independently.

\paragraph{Model-scale heterogeneity.}
This setting retains five ANN clients and five SNN clients but assigns
different channel-width multipliers within each group:
\begin{equation}
\begin{aligned}
\boldsymbol{\rho}_{\mathrm{ann}}
&=
(0.25,\,0.50,\,1.00,\,0.25,\,0.50),\\
\boldsymbol{\rho}_{\mathrm{snn}}
&=
(0.25,\,0.50,\,1.00,\,0.25,\,0.50).
\end{aligned}
\label{eq:app_scale_assignment}
\end{equation}
Each multiplier uniformly scales the channels of its ResNet-18 family
while preserving the residual depth and model type. Client order, data
shards, and local training budgets remain unchanged. The communicated
Bridge has the same size for all clients, independent of private-backbone
width. This setting is evaluated on CIFAR-10 and CIFAR-100.

\paragraph{Architecture heterogeneity.}
The five architectures, in fixed client order, are ResNet-18,
VGG-11, VGG-9, ResNet-18, and MobileNet. ANN clients use their
continuous implementations, while SNN clients use the corresponding
spiking implementations. This setting therefore includes ResNet, VGG,
and MobileNet families in both client groups. The client-to-architecture
assignment, data shards, and initialization seeds are fixed across
methods. No correspondence between intermediate layers of different
architectures is assumed. A baseline is reported only when its original
formulation provides a valid heterogeneous-architecture mapping, without
tensor padding, truncation, or an unreported conversion network. This
setting is evaluated on CIFAR-10 and CIFAR-100.

\paragraph{Client-composition heterogeneity.}
The total number of clients remains ten, while the number of SNN clients
varies from one to nine:
\begin{equation}
N_{\mathrm{snn}}
\in
\{1,\ldots,9\},
\qquad
N_{\mathrm{ann}}
=
10-N_{\mathrm{snn}}.
\label{eq:app_client_composition}
\end{equation}
For every composition, the ordered data shards and Dirichlet partition
remain fixed. Only the predefined client model types are changed, and
the same assignment is used by all methods. This experiment is conducted
on CIFAR-100.

Let $\operatorname{Acc}_{\mathrm{ann}}$ and
$\operatorname{Acc}_{\mathrm{snn}}$ denote the mean accuracies within the
two client groups. The reported overall accuracy is the mean over all ten
clients:
\begin{equation}
\begin{aligned}
\operatorname{Acc}_{\mathrm{all}}
&=
\frac{1}{10}
\sum_{i=1}^{10}
\operatorname{Acc}_i\\
&=
\frac{
N_{\mathrm{ann}}\operatorname{Acc}_{\mathrm{ann}}
+
N_{\mathrm{snn}}\operatorname{Acc}_{\mathrm{snn}}
}{10}.
\end{aligned}
\label{eq:app_ratio_accuracy}
\end{equation}
Therefore, unequal client compositions are weighted by their actual
client counts rather than by an unweighted average of the two group
means.

\subsection{Necessity Experiment}
\label{app:necessity_protocol}

The necessity experiment is conducted on CIFAR-100 with ten clients and
Dirichlet non-IID partitioning at $\alpha=0.1$. Homogeneous controls
contain ten ANN, SNN, or binary-ANN clients. Mixed controls contain five
full-precision ANN clients and five clients of the compared model type.
The ANN-to-ANN control connects two five-client ANN groups through the
same Bridge schedule. For SNN experiments, only the number of time steps
$T\in\{2,4,8\}$ is varied. All other data partitions, model capacities,
and training budgets remain fixed.

The reported computation is the average cost of a single-example forward
pass through the models used by each configuration. It does not include
backpropagation, optimizer updates, repeated local epochs, measured
latency, or hardware energy.

For convolutional layer $\ell$, define its dense multiply-accumulate
count as
\begin{equation}
M_\ell
=
H_\ell W_\ell C_\ell^{\mathrm{out}}
\frac{C_\ell^{\mathrm{in}}}{g_\ell}
K_{\ell,h}K_{\ell,w},
\label{eq:app_conv_macs}
\end{equation}
where $H_\ell$ and $W_\ell$ are the output dimensions,
$C_\ell^{\mathrm{in}}$ and $C_\ell^{\mathrm{out}}$ are the channel
numbers, $g_\ell$ is the number of groups, and
$K_{\ell,h}\times K_{\ell,w}$ is the kernel size. Counting one
multiplication and one addition as two floating-point operations gives
\begin{equation}
C_{\mathrm{F},\ell}
=
2M_\ell.
\label{eq:app_dense_flops}
\end{equation}
For a linear layer,
$M_\ell=d_\ell^{\mathrm{in}}d_\ell^{\mathrm{out}}$ and
$C_{\mathrm{F},\ell}=2M_\ell$. Other dense operations, such as
normalization, pooling, and neuron updates, are included according to
their executed scalar-operation counts when present.

For a spike-driven layer, let $\rho_{\ell-1}^{t}$ denote the measured
fraction of nonzero input spikes at time step $t$. Its synaptic
accumulation count is
\begin{equation}
C_{\mathrm{S},\ell}
=
M_\ell
\sum_{t=1}^{T}
\rho_{\ell-1}^{t}.
\label{eq:app_snn_sops}
\end{equation}
The firing rates are measured from the trained model on the fixed
evaluation set. Spike-driven accumulations are reported as SOPs, while
any dense operations retained by the SNN implementation are reported as
FLOPs. The two operation types are not converted into one another using
an assumed hardware-dependent energy ratio.

Let $C_{\mathrm{F},i}$ and $C_{\mathrm{S},i}$ denote the summed FLOPs and
SOPs of client $i$, and let $C_{\mathrm{F},B}$ denote one Bridge forward
pass. The computation reported for a configuration is
\begin{equation}
\begin{aligned}
C_{\mathrm{F}}^{\mathrm{cfg}}
&=
\frac{1}{10}
\sum_{i=1}^{10}
\left(
C_{\mathrm{F},i}
+
\delta_i C_{\mathrm{F},B}
\right),\\
C_{\mathrm{S}}^{\mathrm{cfg}}
&=
\frac{1}{10}
\sum_{i=1}^{10}
C_{\mathrm{S},i},
\end{aligned}
\label{eq:app_configuration_cost}
\end{equation}
where $\delta_i=1$ when client $i$ uses the Bridge and $0$ otherwise.
Thus, a Bridge-based row includes one private-model forward pass and one
Bridge forward pass per participating client. Both totals are divided by
$10^6$ to obtain the reported million-operation units.

Binary convolutions are reported separately as XNOR-popcount operations.
They are not converted to FLOPs or SOPs, ensuring that the table remains
independent of a particular hardware or energy model.

\subsection{Ablation Experiment}
\label{app:ablation_protocol}

The ablation experiment is conducted on CIFAR-100 with Dirichlet
non-IID partitioning at $\alpha=0.1$. It uses five ResNet-18 ANN clients,
five Spiking ResNet-18 clients, and $T=4$. All variants use the same data
partition, initialization seeds, communication rounds, local epochs,
optimizer, learning-rate schedule, and checkpoint rule. Local supervised
training remains active for every client.

For the ANN $\rightarrow$ Bridge $\rightarrow$ SNN variant, ANN clients
inject local knowledge into the Bridge and SNN clients extract knowledge
from it. SNN injection and ANN extraction are disabled. The reverse
variant retains SNN injection and ANN extraction while disabling the
opposite direction. These two variants isolate the contribution of each
cross-type knowledge-transfer direction.

Without the personalized Bridge head, the client-specific heads are
replaced by a single head that is communicated and aggregated together
with the Bridge body. Without the Pseudo-Spike Interface, the
pseudo-spike path, rate-matching loss, and Pseudo-Spike Polarization
Regularization are removed. The continuous Bridge path and bidirectional
logit distillation remain unchanged. Table~\ref{tab:ablation}
therefore compares each variant with the complete AS-FedBridge under an
identical training protocol.

\subsection{Communication and Training Cost}
\label{app:cost_protocol}

\paragraph{Communication cost.}
All communicated values are stored in FP32 unless stated otherwise.
Following the convention used in the tables, one MB denotes $2^{20}$
bytes. If a method transmits $P$ scalar values, its one-way payload is
\begin{equation}
\begin{aligned}
C_{\mathrm{payload}}
&=
\frac{4P}{2^{20}}\ \mathrm{MB},\\
C_{\mathrm{client}}
&=
2C_{\mathrm{payload}},\\
C_{\mathrm{server}}
&=
2N C_{\mathrm{payload}}.
\end{aligned}
\label{eq:app_communication}
\end{equation}
The communication tables report $C_{\mathrm{payload}}$, namely the size
of the object transmitted in one direction. Under symmetric download and
upload, $C_{\mathrm{client}}$ is the complete per-client traffic per
round, while $C_{\mathrm{server}}$ is the total traffic for $N$
participating clients.

For each method, $P$ includes only the tensors actually exchanged in one
round. AS-FedBridge communicates the shared Bridge body $\omega$.
Private backbones, personalized heads, adapters, optimizer states,
membrane potentials, and spike states are excluded because they never
leave the client.

\paragraph{Local training cost.}
Let client $i$ contain $n_i$ training examples and use a maximum batch
size $b$. The number of batches in one local epoch and the size of batch
$u$ are
\begin{equation}
\begin{aligned}
K_i
&=
\left\lceil\frac{n_i}{b}\right\rceil,\\
b_{i,u}
&=
\min\left\{
b,\,
n_i-(u-1)b
\right\}.
\end{aligned}
\label{eq:app_local_batches}
\end{equation}
Thus, the final incomplete batch is evaluated at its actual size rather
than being counted as a full batch.

Let $\operatorname{Fwd}_i(q)$ and
$\operatorname{Bwd}_i(q)$ denote the profiled forward and backward costs
of the private model for a batch of size $q$. The corresponding Bridge
costs are $\operatorname{Fwd}_B(q)$ and
$\operatorname{Bwd}_B(q)$. Backward cost excludes the forward pass
already counted separately. For AS-FedBridge, the per-batch costs of
extraction and injection are
\begin{equation}
\begin{aligned}
c_i^{\mathrm{ext}}(q)
&=
\operatorname{Fwd}_B(q)
+
\operatorname{Fwd}_i(q)
+
\operatorname{Bwd}_i(q),\\
c_i^{\mathrm{inj}}(q)
&=
\operatorname{Fwd}_i(q)
+
\operatorname{Fwd}_B(q)
+
\operatorname{Bwd}_B(q).
\end{aligned}
\label{eq:app_stage_cost}
\end{equation}
During extraction, the Bridge is the frozen teacher and only the private
model is updated. During injection, the private model is frozen and only
the Bridge is updated. The frozen teacher therefore incurs a forward pass
but no backward pass.

If extraction and injection use $E_{\mathrm{ext}}$ and
$E_{\mathrm{inj}}$ local epochs, respectively, the cost of one client
round is
\begin{equation}
\begin{aligned}
C_i^{\mathrm{ext}}
&=
E_{\mathrm{ext}}
\sum_{u=1}^{K_i}
c_i^{\mathrm{ext}}(b_{i,u}),\\
C_i^{\mathrm{inj}}
&=
E_{\mathrm{inj}}
\sum_{u=1}^{K_i}
c_i^{\mathrm{inj}}(b_{i,u}),\\
C_i^{\mathrm{round}}
&=
C_i^{\mathrm{ext}}
+
C_i^{\mathrm{inj}}.
\end{aligned}
\label{eq:app_training_cost}
\end{equation}
The reported per-example local-training cost is averaged over clients as
\begin{equation}
C_{\mathrm{train}}
=
\frac{1}{N}
\sum_{i=1}^{N}
\frac{C_i^{\mathrm{round}}}{n_i}.
\label{eq:app_training_average}
\end{equation}
A shard-level total instead reports $C_i^{\mathrm{round}}$ without
division by $n_i$.

For each baseline, the same calculation is applied to the computation
graph it actually executes, including any public model, messenger,
prototype module, or additional local objective. Forward and backward
costs are obtained using the same profiler for all methods rather than
approximating backward computation by a fixed multiple of forward
computation.

Dense operations are reported as FLOPs. Spike-driven synaptic
accumulations are summed over all time steps and reported as SOPs, while
dense Bridge or messenger operations remain in FLOPs. Loss evaluation
and optimizer scalar updates are excluded consistently for all methods.
These values measure operation counts rather than wall-clock latency or
hardware energy.

\section{Method Details}
\label{app:theory}

\begin{algorithm}[t]
\caption{Client Extraction and Injection}
\label{alg:app_client}
\small
\begin{algorithmic}[1]
\REQUIRE Client $i$, received body $\omega^r$, private parameters
$\theta_i,\psi_i$, local epochs $E_{\mathrm{ext}},E_{\mathrm{inj}}$
\ENSURE Updated private parameters and uploaded body $\omega_i^{r+1}$

\STATE Set $\omega_i\leftarrow\omega^r$ and attach private head $\psi_i$

\STATE \textit{Extraction: update the private backbone}
\STATE Freeze $\omega_i,\psi_i$ and unfreeze $\theta_i$
\FOR{$e=1,\ldots,E_{\mathrm{ext}}$}
  \FOR{mini-batch $(x,y)\subset\mathcal D_i$}
    \STATE Compute detached Bridge logits $Z_B$
    \IF{$i\in\mathcal C_{\mathrm{ann}}$}
      \STATE Compute $Z_i$ and update $\theta_i$ using CE and KD
    \ELSE
      \STATE Reset SNN states and compute $\{Z_i^t\}_{t=1}^{T}$
      \STATE Update $\theta_i$ using TET and NLD-based KD
    \ENDIF
  \ENDFOR
\ENDFOR

\STATE \textit{Injection: update the local Bridge}
\STATE Freeze $\theta_i$ and unfreeze $\omega_i,\psi_i$
\FOR{$e=1,\ldots,E_{\mathrm{inj}}$}
  \FOR{mini-batch $(x,y)\subset\mathcal D_i$}
    \IF{$i\in\mathcal C_{\mathrm{ann}}$}
      \STATE Compute detached $Z_i$ and Bridge logits $Z_B$
      \STATE Update $\omega_i,\psi_i$ using CE, KD, and proximal losses
    \ELSE
      \STATE Reset SNN states and compute detached $\bar Z_i,r_i$
      \STATE Compute Bridge logits $Z_B$ and rates $\widetilde r_B$
      \STATE Update $\omega_i,\psi_i$ using CE, KD, rate,
      PSPR, and proximal losses
    \ENDIF
  \ENDFOR
\ENDFOR

\STATE Retain $\theta_i,\psi_i$ and optimizer states locally
\STATE Upload only $\omega_i$ to the server
\STATE \textbf{return} $\omega_i^{r+1}\leftarrow\omega_i$
\end{algorithmic}
\end{algorithm}

\subsection{Theoretical Role and Gradient of PSPR}
\label{app:pspr}

Pseudo-Spike Polarization Regularization (PSPR) is applied to the
continuous rate variables before quantization. For selected interfaces
$\mathcal P$, let $\mu_\ell$ and $\sigma_\ell$ denote the channel-wise
mean and standard deviation of $r_B^\ell$. With
$(u)_+=\max(u,0)$, PSPR is
\begin{equation}
\begin{aligned}
\mathcal L_{\mathrm{PSPR}}
&=
\frac{1}{|\mathcal P|}
\sum_{\ell\in\mathcal P}
\mathbb E
\left[
\ell_{\mu,\ell}
+
\lambda_{\mathrm{var}}\ell_{\sigma,\ell}
\right],\\
\ell_{\mu,\ell}
&=
\left(
|\mu_\ell-V_{\mathrm{th}}|
-k_d\sigma_\ell
\right)_+^2,\\
\ell_{\sigma,\ell}
&=
\left(
\sigma_{\min}-\sigma_\ell
\right)_+^2.
\end{aligned}
\label{eq:app_pspr}
\end{equation}
A zero PSPR penalty at port $\ell$ requires
\begin{equation}
|\mu_\ell-V_{\mathrm{th}}|
\le
k_d\sigma_\ell,
\qquad
\sigma_\ell
\ge
\sigma_{\min}.
\label{eq:app_pspr_zero_set}
\end{equation}
The first condition keeps the threshold within a controlled
mean-deviation range, while the second discourages vanishing
pre-quantization variance. PSPR therefore reduces moment collapse but
does not determine the complete activation distribution or guarantee
occupancy of every discrete firing-rate level.

The three interface components have complementary roles.
$Q_T$ restricts the forward values to
$\mathcal R_T=\{0,1/T,\ldots,1\}$,
$\mathcal L_{\mathrm{rate}}$ aligns individual Bridge representations
with observed SNN firing rates, and $\mathcal L_{\mathrm{PSPR}}$
maintains variation before quantization. Consequently, the multi-peak
distributions observed in the experiments arise jointly from data,
rate supervision, and quantization rather than from PSPR alone.

To show how this objective remains trainable, consider one scalar
activation:
\begin{equation}
\begin{aligned}
u
&=
\frac{\operatorname{ReLU}(a)}{s},\\
r
&=
\operatorname{clip}_{[0,1]}(u),\\
\widetilde r
&=
r+\operatorname{sg}
\left(
Q_T(r)-r
\right),
\end{aligned}
\label{eq:app_gradient_chain}
\end{equation}
where $s>0$ is the learned scale. The straight-through estimator gives
$\partial\widetilde r/\partial r=1$. Under the ordinary clipping
derivative, the rate-branch gradient can be written as
\begin{equation}
\begin{aligned}
\frac{\partial\mathcal L_{\mathrm{port}}}{\partial a}
&=
\frac{
\mathbf 1[a>0]\mathbf 1[0<u<1]
}{s}
\,\Delta_r,\\
\Delta_r
&=
\alpha_{\mathrm{rate}}
\frac{\partial\mathcal L_{\mathrm{rate}}}
{\partial\widetilde r}
+
\alpha_{\mathrm{pspr}}
\frac{\partial\mathcal L_{\mathrm{PSPR}}}
{\partial r}.
\end{aligned}
\label{eq:app_port_gradient}
\end{equation}
Thus, quantization does not block the surrogate gradient. The local
rate branch becomes inactive only when the ReLU output is negative or
the normalized value is clipped at a boundary. Even in this case, the
continuous Bridge classification path still receives cross-entropy and
logit-distillation gradients, allowing subsequent encoder updates to
move activations back into the valid interval. This provides a practical
recovery path, although it does not constitute a formal guarantee against
complete saturation.

\subsection{Complete Training Procedure}
\label{app:algorithm}

Algorithm~\ref{alg:app_client} details the two-stage local update, while
Algorithm~\ref{alg:app_server} summarizes server aggregation. The
server communicates and aggregates only the shared Bridge body.

\begin{algorithm}[t]
\caption{Server Aggregation in AS-FedBridge}
\label{alg:app_server}
\small
\begin{algorithmic}[1]
\REQUIRE Initial Bridge body $\omega^0$, client set
$\{(\mathcal D_i,\theta_i,\psi_i)\}_{i=1}^{N}$, rounds $R$
\ENSURE Aggregated Bridge body $\omega^R$

\FOR{$r=0,\ldots,R-1$}
  \STATE Select participating clients $\mathcal S_r$
  \STATE Broadcast $\omega^r$ to all $i\in\mathcal S_r$
  \FOR{each client $i\in\mathcal S_r$}
    \STATE $\omega_i^{r+1}\leftarrow
    \operatorname{ClientUpdate}(i,\omega^r)$
  \ENDFOR
  \STATE $\displaystyle
  \omega^{r+1}\leftarrow
  \sum_{i\in\mathcal S_r}
  \frac{|\mathcal D_i|}
  {\sum_{j\in\mathcal S_r}|\mathcal D_j|}
  \omega_i^{r+1}$
\ENDFOR

\STATE \textbf{return} $\omega^R$
\end{algorithmic}
\end{algorithm}

Extraction and injection use the same local data but optimize disjoint
parameter sets. The frozen Bridge first teaches the private backbone,
after which the updated backbone becomes the teacher of the local Bridge.
Only the shared Bridge body is uploaded and aggregated. At inference,
each client uses its private ANN logits $Z_i$ or temporally averaged SNN
logits $\bar Z_i$, without invoking the Bridge.
\end{document}